%% file: arxiv.tex
\documentclass{article} 
\usepackage{arxiv,times}

\input{math_commands.tex}

\usepackage{url}
\usepackage{algorithm}
\usepackage{algorithmic}
\usepackage{amssymb}
\usepackage{xspace}
\usepackage{graphicx}
\usepackage{multirow}
\usepackage[table]{xcolor}
\usepackage{colortbl}
\usepackage{makecell}
\usepackage{booktabs}
\usepackage{wrapfig}
\usepackage{longtable}
\usepackage{float}
\usepackage{color}
\definecolor{citecolor}{HTML}{FA7A94}
\definecolor{nvidiaGreen}{RGB}{118,185,0}
\usepackage[colorlinks=true,
            linkcolor=citecolor,
            citecolor=citecolor,
            urlcolor=citecolor]{hyperref}
\usepackage{svg}
\usepackage{cleveref}
\usepackage[misc]{ifsym}

\title{Self-Improving Vision-Language-Action\\Models with Data Generation via Residual RL}

\author{Wenli Xiao$^{1\,2\,\dag}$, \ Haotian Lin$^{2\,\dag}$,  Andy Peng$^{1\,3}$, Haoru Xue$^{1\,3}$, Tairan He$^{1\,2}$, Yuqi Xie$^{1}$,\\
    \textbf{Fengyuan Hu$^{1}$, Jimmy Wu$^{1}$, Zhengyi Luo$^{1}$, Linxi ``Jim'' Fan$^{1}$, Guanya Shi$^{2}$, Yuke Zhu$^{1\,4}$} \\
    $^{1}$NVIDIA, $^{2}$CMU, $^{3}$UC Berkeley, $^{4}$UT Austin, $^\dag$Equal Contributions \\
}

\renewcommand{\arraystretch}{1.4}
\newcommand{\pagelink}{https://wenlixiao.com/self-improve-VLA-PLD}

\newcommand{\su}[2]{\makecell{#1 \\ #2}}
\definecolor{cmuRed}{HTML}{C41230}
\newcommand{\method}{\texttt{\textbf{\textcolor{nvidiaGreen}{PLD}}}\xspace}
\newcommand{\eg}{e.g.,\ }
\newcommand{\ie}{i.e.,\ }

\arxiv 
\begin{document}

\maketitle

\begin{abstract}

\begin{figure*}[!b]
    \centering
    \includegraphics[width=1.0\linewidth]{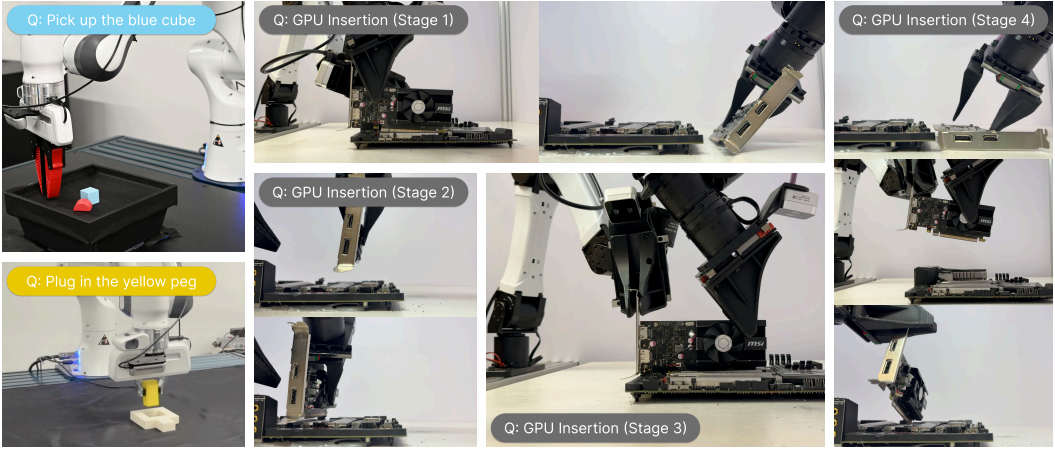}
    \caption{We demonstrate the performance of \method on several real-world challenging manipulation tasks. The robot successfully picks up diverse objects and conduct peg insertion for Franka arm. Besides, we also deploy \method on YAM bi-manual settings, showing \method policy continuously perform a continuous 1-hour cycle of GPU insertion and unplugging \emph{without human resets}. (Videos at~\url{\pagelink})}
    \label{fig: real-world-setup} 
\end{figure*}

\input{sections/0_abstract}
\end{abstract}

\input{sections/1_introduction}

\input{sections/3_preliminaries}

\input{sections/4_methods}

\input{sections/5_experiments_arxiv}

\input{sections/2_related_works}

\input{sections/6_conclusions}

\input{sections/8_acknowledgement}

\bibliography{arxiv}
\bibliographystyle{arxiv}

\newpage
\appendix

\input{sections/7_appendix_arxiv}

\end{document}

%% file: math_commands.tex
\usepackage{amsmath,amsfonts,bm}

\def\eqref#1{equation~\ref{#1}}

\def\1{\bm{1}}

\DeclareMathAlphabet{\mathsfit}{\encodingdefault}{\sfdefault}{m}{sl}
\SetMathAlphabet{\mathsfit}{bold}{\encodingdefault}{\sfdefault}{bx}{n}



%% file: sections/0_abstract.tex
Supervised fine-tuning (SFT) has become the de facto post-training strategy for large vision-language-action (VLA) models, but its reliance on costly human demonstrations limits scalability and generalization. We propose Probe, Learn, Distill (\method), a three-stage plug-and-play framework that improves VLAs through residual reinforcement learning (RL) and distribution-aware data collection. In Stage~1 (\emph{specialist acquisition}), we freeze the VLA backbone and train lightweight residual actors via off-policy RL. These specialists take over in states where the base policy fails, thereby probing failure regions of the VLA generalist. In Stage~2 (\emph{data collection}), we employ a hybrid rollout scheme that biases residual interventions toward states frequently visited by the base policy, aligning collected trajectories with the generalist’s deployment distribution while capturing recovery behaviors. In Stage~3 (\emph{fine-tuning}), these curated trajectories are distilled back into the generalist with standard SFT, applicable to both flow-matching and autoregressive heads. We evaluate \method across diverse settings: it achieves a near-saturated 99\% task success rate on the LIBERO benchmark, delivers over 50\% performance gains in SimplerEnv, and demonstrates a 100\% success rate on real-world Franka arm and YAM arm dexterous manipulation tasks. We further provide ablations showing that residual policy probing and distribution-aware replay are key to collecting deployment-aligned data that improves VLAs’ capabilities on both seen and unseen tasks. Our results demonstrate that RL-generated, policy-aligned data can surpass teleoperation-only demonstrations, offering a scalable path toward self-improving VLA models.


%% file: sections/1_introduction.tex
\section{Introduction}
\begin{figure}
    \label{fig:cross-task}
    \centering
    \includegraphics[width=0.95\linewidth]{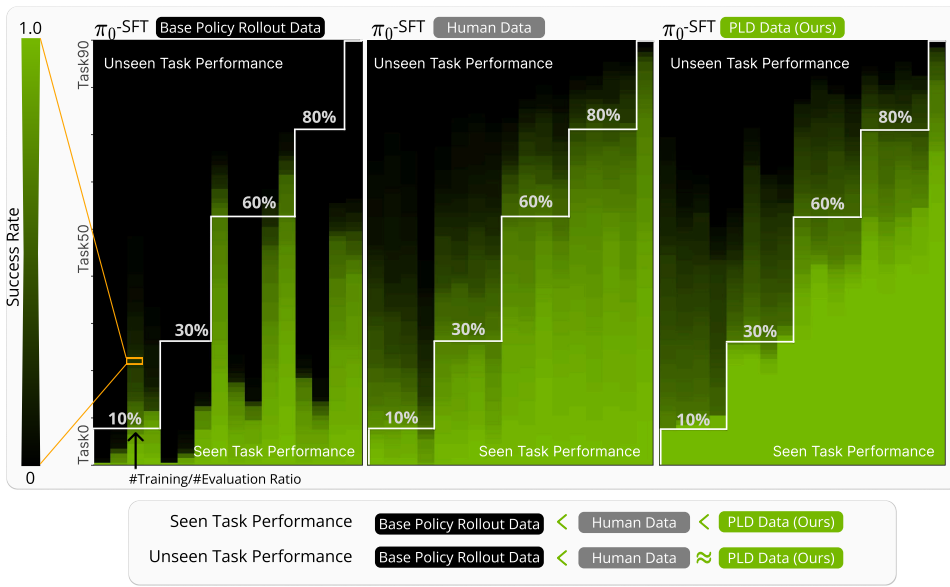}
    \caption{\textbf{Synergetic effect of \method\ data.} We fine-tune $\pi_0$ on subsets of LIBERO-90 with varying \textbf{task coverage ratios}, where each ratio (10–80\%) indicates the fraction of distinct task instances included in training relative to the full 90-task distribution. For each ratio, we randomly sample 4 disjoint subsets of tasks and report the averaged results. The x-axis thus represents the degree of task coverage (not the number of trajectories), while the evaluation is always conducted on all 90 tasks. We compare different data formulations: \method\ data yields the highest in-distribution performance while retaining the cross-task generalization property of high-quality human data. It further enables modest-level zero-shot transfer even when trained on only 10\% of tasks (24.4\% SR on unseen tasks), whereas the VLA fine-tuned on base-policy rollout data (0-1 REINFORCE) underperforms and fails to generalize. (Success rate numbers are reported in~\Cref{tab:libero-90-seen-to-unseen}.)}
    \label{fig:seentask-to-unseentask}
\end{figure}

Supervised fine-tuning (SFT) has become the standard post-training paradigm for large language models (LLMs): after broad pre-training, models are adapted to downstream applications by training on curated instruction–response pairs, yielding many improvements in language following, safety, and generalization~\citep{ouyang2022traininglanguagemodelsfollow,geminiteam2025geminifamilyhighlycapable}. Inspired by these successes, the same recipe is now being applied to robot foundation models, particularly vision-language-action (VLA) policies, where large, heterogeneous robotics and vision-language datasets provide the base initialization, and SFT specializes models to specific tasks and embodiments~\citep{o2024open,team2024octo,kim2024openvla,openvla-oft,black2024pi0,bjorck2025gr00t}. However, transferring this paradigm from language to robotics is a unique challenge. Collecting high-quality robot demonstrations is both costly and labor-intensive, making large-scale datasets much harder to obtain. Even when such data are available, they are often collected through teleoperation pipelines that are \emph{decoupled} from the deployed VLA policy, leaving critical coverage gaps: human operators must manually anticipate and correct failure modes, but their demonstrations rarely reflect the actual distribution of states the policy will encounter at deployment. As a result, while SFT reliably improves performance on the tasks it is trained on, much less is understood about whether these gains transfer to new tasks and environments.  

These challenges raise the following question: Can VLA models improve themselves using RL-curated data with minimal human effort? Specifically, can this self-curated training match or surpass fine-tuning on human-expert (oracle) teleoperation data, both in-distribution and out-of-distribution? Our central observation is that data collection should \textit{not} be agnostic to the base policy: the data-collecting policy and the generalist must \emph{interact}, so that exploration leverages the generalist’s prior knowledge and collected data remain aligned with its trajectory distribution. A natural way to instantiate this idea is to employ reinforcement learning (RL) to acquire task-specific specialists that guide data collection. However, applying RL in this setting is hindered by two key challenges. Sparse reward signals in language-conditioned manipulation tasks render RL unstable and sample-inefficient. Moreover, training task-specific experts independently from the generalist introduces distributional mismatch, and once these experts converge, their behavior often lacks the diversity needed to provide robust coverage for SFT.

Motivated by these challenges, we introduce \method, a three-stage post-training pipeline. \textbf{Stage~1: Online specialist acquisition.} We \emph{freeze} the VLA backbone and train several lightweight \emph{residual} actors for multiple tasks via sample-efficient off-policy RL, enabling them to “take over” the base policy at arbitrary states and achieve above 99\% task success. \textbf{Stage~2: Automatic data collection.} We propose a hybrid rollout scheme that biases residual takeovers toward states frequently visited by the base model, mitigating distribution shift while capturing recovery behaviors. \textbf{Stage~3: Supervised fine-tuning.} The collected data for multiple tasks are distilled back into the base model through SFT, a process agnostic to VLA architectures, supporting both flow-matching and autoregressive action heads~\citep{black2024pi0,kim2024openvla}. An overview of our pipeline can be found in ~\Cref{fig:overview}. With \method, we can efficiently acquire task-specific RL experts through VLA-guided exploration. Consequently, the VLA further improves using the \method data, achieving performance above 99\% on the LIBERO benchmark.

This paper makes the following contributions: 1) Autonomous post-training recipe. We propose a post-training pipeline that enables VLA models to improve autonomously without relying on additional oracle demonstrations. Our method achieves near-saturated 99\% success rates on the LIBERO benchmark, and delivers over 50\% performance gains in SimplerEnv, underscoring both its effectiveness on seen tasks and its ability to generalize to unseen ones. 2) Systematic study of RL-generated data. We analyze the key components of automatic data collection most beneficial for SFT, and conduct extensive experiments in simulation and on real robot hardware to examine how \emph{RL-generated data} influences generalization to unseen tasks. 3) Comprehensive empirical validation. We provide large-scale ablations of our design choices. Besides, we showcase $>$99\% success rate on Franka arm and YAM arm dexterous manipulation tasks. Achieving continuous GPU insertion and unplugging operating for 1 hour without human intervention, offering potential for data-efficient post-training of robot foundation models.

%% file: sections/3_preliminaries.tex
\section{Preliminaries}
\label{sec: preliminaries}

\subsection{Task formulation}
We study language-conditioned manipulation with \emph{sparse binary rewards} using \emph{Vision--Language--Action} (VLA) models as the base policy class. We assume a partially observed control process with horizon $T$, where an episode terminates and resets on task success with a restricted time limit. After each episode, a reward $r\in\{0,1\}$ is assigned. Let $g$ denote the language prompt of goal specification, and let $o_t$ denote partial observations comprising robot proprioception (\eg joint angle) and RGB images input. The policy consumes $(o_t,g)$ and outputs a 7-DoF action (6-DoF delta pose and 1-DoF continuous gripper command), which we express as $a_t \;=\; D_\phi\!\big(h_\theta(o_t,g)\big)$, 
where $h_\theta$ is a vision--language backbone and $D_\phi$ is an action head. Consistent with recent VLA models, $D_\phi$ is instantiated by one of three common families: (i) a \emph{diffusion} or \emph{flow-based} action head for continuous control~\citep{team2024octo, black2024pi0}, or (ii) a \emph{discrete action tokenizer} for autoregressive decoding~\citep{kim2024openvla, pertsch2025fast}. We aim to maximize success rate $\bar{\Sigma r}$ by tuning $\phi$ and $\theta$. 

\subsection{Supervised Fine-Tuning}
Given a VLA policy and a demonstration dataset
$\mathcal{D}=\{(o_t,g_t,a_t)\}$ of observations $o_t$,
goal specifications $g_t$, and expert actions $a_t$, SFT adapts the policy by maximizing the conditional action likelihood. Letting $x_t=(o_t,g_t)$, the canonical objective is the behavior cloning (BC) loss.
In contemporary VLA systems, the loss instantiation depends on the action head architecture. Auto-regressive/token heads~\citep{kim2024openvla, pertsch2025fast} train with sequence NLL over
action tokens $u_{1:K}$:
\[
\mathcal{L}_{\text{AR}}(\theta)
= - \,\mathbb{E}_{k \sim [K]}\!\left[\log p_{\theta}\!\big(u_k \mid u_{<k}, x\big)\right],
\]

With recent work improving efficiency via action chunking and parallel decoding, and a continuous action parameterization trained by an $\ell_1$ regression
objective~\citep{openvla-oft}. 
Diffusion heads model a conditional denoising
process for actions and train via score-matching MSE:
\[
\mathcal{L}_{\text{diff}}(\theta)
= \mathbb{E}_{t,\epsilon,(x,a)}\!\left[\big\|\epsilon - \epsilon_{\theta}(a_t^{(\text{noisy})}, x, t)\big\|_2^2\right],
\]
enabling iterative sampling at inference~\citep{team2024octo,chi2024diffusionpolicy}. 
Flow-matching heads learn a continuous velocity field to
transport a prior to the action distribution, trained with an $L_2$ flow-matching loss,
and are often paired with VLM backbones for semantically grounded control~\citep{black2024pi0, intelligence2025pi05}. 
Across these heads, SFT remains the standard mechanism to specialize a generalist
policies to new embodiments and tasks using modest labeled robot data~\citep{kim2024openvla, openvla-oft}.

\subsection{Goal-Conditioned RL}
We model continuous control as an MDP~\citep{bellman1957MDP}
$\mathcal{M}=(\mathcal{S},\mathcal{A},\rho,\rho_0,r,\gamma)$ with state space
$\mathcal{S}$, action space $\mathcal{A}$, transition dynamics
$\rho(s'\mid s,a)$, initial-state distribution $\rho_0$, reward function $r$,
and discount $\gamma\in(0,1]$. In \emph{goal-conditioned} settings, each task
is specified by a goal variable $g\in\mathcal{G}$ drawn from $p(g)$; the reward
becomes goal-dependent $r:\mathcal{S}\times\mathcal{A}\times\mathcal{G}\to\mathbb{R}$,
and the policy is $\pi:\mathcal{S}\times\mathcal{G}\to\Delta(\mathcal{A})$,
written $\pi(a\mid s,g)$. It is convenient to view GCRL as an augmented MDP on
$\mathcal{S}\times\mathcal{G}$ with stationary goals:
\[
\tilde{\rho}\big((s',g)\mid(s,g),a\big)=\rho(s'\mid s,a)\cdot \mathbf{1}\{g'=g\}.
\]
Under the infinite-horizon setting, the RL objective is
\begin{equation}
\label{eq:gcrl-objective}
J(\pi)
= \mathbb{E}_{g \sim p(g)}\;
  \mathbb{E}_{s_0 \sim \rho_0,\;
             a_t \sim \pi(\cdot \mid s_t,g),\;
             s_{t+1} \sim \rho(\cdot \mid s_t,a_t)}
  [\sum\nolimits_{t=0}^{\infty}
 \gamma^t\, r(s_t,a_t,g)].
\end{equation}

In this paper, we consider a sparse binary reward setting, \ie $r(s,a,g)=\mathbf{1}\!\left[d\big(\phi(s),g\big)\le \varepsilon\right]$ defined via a success
predicate over a goal-relevant representation $\phi(s)$, a metric $d$, and tolerance $\varepsilon>0$.

%% file: sections/4_methods.tex
\section{Methods}
\label{sec: methods}

\begin{figure}
    \centering
    \includegraphics[width=1\linewidth]{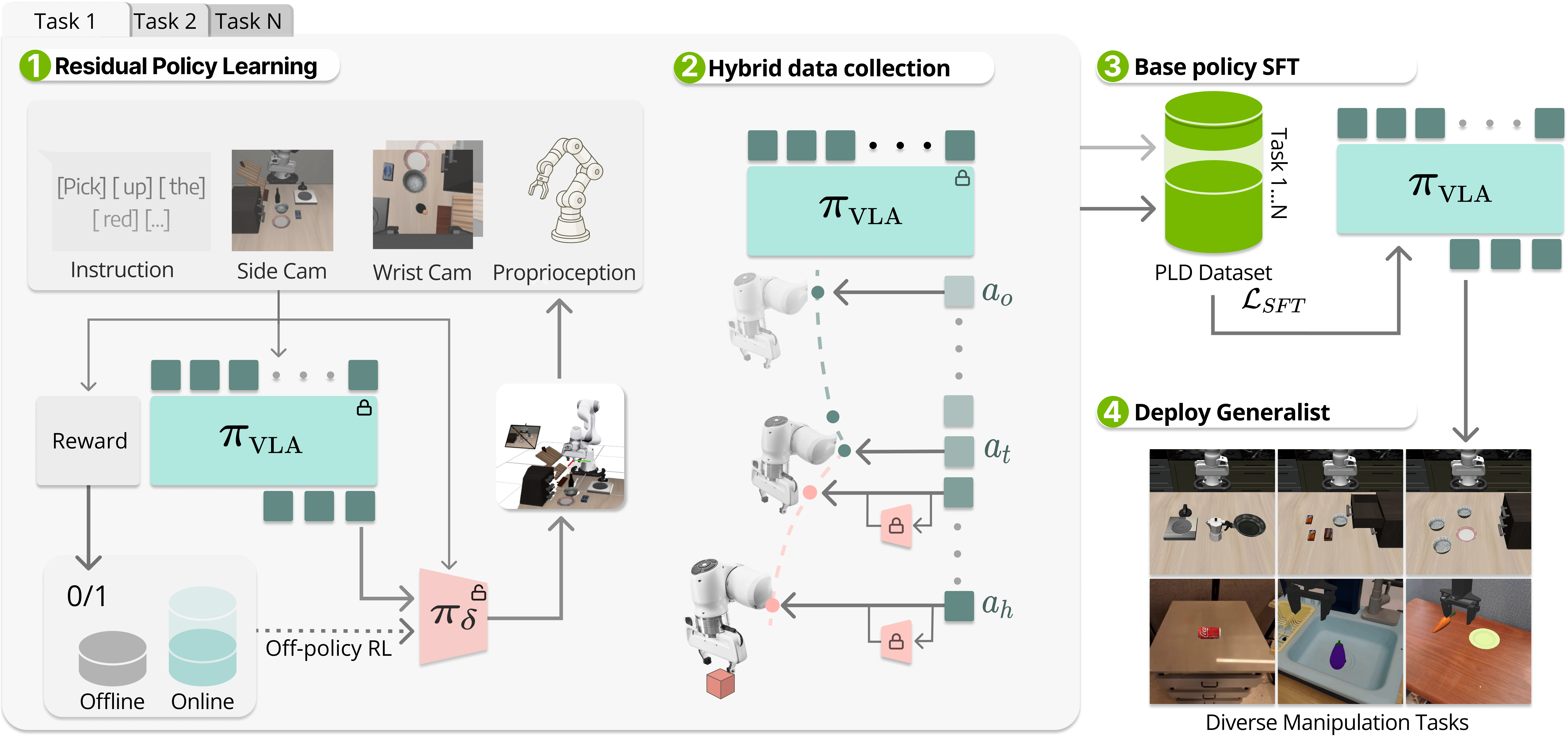}
    \caption{\textbf{An overview of \method.} Our pipeline consists of three stages: 1) learning specialist residual policy for each task via online off-policy RL, with efficient exploration guided by a frozen VLA generalist; 2) Automatic generation of hybrid trajectories by having the VLA rollout for the first $t$ steps and let the specialist takeover to generate recovery data; 3) Supervised fine-tuning using collected multi-task \method data; 4) Deploy the fine-tuned generalist to diverse manipulation tasks in zero-shot.}
    \label{fig:overview}
\end{figure}

\paragraph{Method Overview}
We study the synergy between data produced by our method when a modest \emph{generalist} VLA serves as the policy prior. The premise is that, if we exploit the base policy’s prior correctly, it can both \emph{solve hard tasks quickly} and \emph{explore efficiently}. While recent work explores direct RL fine-tuning of large VLA~\citep{mark2024PARL, dong2025expo}, such formulas can be resource-intensive even for single-task tuning: \eg OpenVLA-OFT requires per-GPU memory up to $\sim$62.5\ GB for LIBERO training at a batch size of~8~\citep{openvla-oft}. Meanwhile, it remains unclear whether these approaches scale gracefully to multi-task fine-tuning under heterogeneous setups. We therefore opt for a \emph{decoupled} pipeline. We freeze the base policy $\pi_b$ and learn a lightweight residual action policy $\pi_{\delta}$ with sample-efficient off-policy RL (Gaussian policy parameterization). We then \emph{collect expert data} by letting the residual ``take over’’ after specified steps of ``base policy probing’’. Finally, we \emph{distill} these skills back into the base model via SFT and deploy the generalist on diverse manipulation tasks. We provide the overview of \method in~\Cref{fig:overview}.


\subsection{Data Efficient RL via Policy Prior Warm-start}

Building upon the previous success of sample-efficient RL with prior data~\citep{RLPD}, we consider an off-policy actor-critic framework and maintain two separate buffers for offline and online experience replay. We first fill the offline buffer with successful rollouts $\mathcal{B}_{offline} = \{\tau_1, \tau_2, \dots\}$ from the base policy $\pi_{b}$. This process serves as an importance sampling to preserve only the successful attempts. During training, the offline and online experiences will be replayed symmetrically; for example, mini-batches consist of equal samples from both buffers, ensuring that the value function is constantly trained on high-value state-action pairs. 

In practice, we train a task-specific residual action module $\pi_{\delta}(\cdot | s, a_b)$ conditioned on $a_b \sim \pi_b$. We use $\pi_{\delta}$ to explore near the base policy behavior, actively searching for more optimal solutions guided by the Q-function. To modulate exploration and avoid deviating drastically from $\pi_{b}$ during the initial phase, the delta action's magnitude is scaled down to $[-\xi, \xi]$, where $\xi \in [0, 1]$ is tuned by a scheduler. This design choice is two-fold: First, although unable to perfectly generalize to an unseen manipulation task or scenario, the base policy can make reasonable attempts to solve the task, serving as a useful initialization for exploration. Moreover, directly training the expressive foundation policy (\eg flow action heads) to maximize the Q-value can be extremely difficult \citep{mark2024PARL}. In contrast, a residual Gaussian policy can be easily trained through any off-the-shelf off-policy RL algorithm.

Alongside $\pi_{\delta}$, action value function $Q^{\bar{
\pi}}$ acquired through policy iteration and TD-learning~\citep{sutton2018reinforcement} as in~\Cref{eqn: Q def}, where $\bar{\pi}(\cdot | s)$ is the combined policy.
\begin{equation}
    \label{eqn: Q def}
    Q^{\bar{\pi}}(s_t, \bar{a}_t) \leftarrow r(s, a) + \gamma \mathbb{E}_{s_{t+1}\sim p(\cdot | s_t, \bar{a}_t)} [Q_{target}^{\bar{\pi}}(s_{t+1}, \bar{a}_{t+1}) ] 
     , \ \bar{a}=a_b+a_\delta
\end{equation}

To stabilize off-policy learning and mitigate forgetting, we introduce a warm-up stage using solely $\pi_b$ for data collection akin to~\citep{WSRL}. Meanwhile, the Q-function is initialized by a conservative objective such as Cal-QL~\citep{cal-QL}. Importantly, we do not explicitly enforce behavior constraints to policy loss, such that the resulting expert \(\bar{\pi}\) is less influenced by either data quality or base policy performance.

\subsection{Bootstrapping RL Specialist for Scalable Data Generation}
We then turn to the question of how to collect demonstration data using RL specialists. Data collected through RL experts is highly optimal, with consistent behavior and nearly no hesitation, demonstrating smooth solutions that finish tasks with a shorter horizon. However, such a narrow distribution of unimodal expert behavior may leave out-of-distribution and failure states underrepresented. Thus, scaling purely expert data may not result in a performance gain, but instead risks the generalist overfitting on these data and harming both robustness and generalization (As discussed in the following section).

To mitigate this issue, we propose a hybrid data collection scheme that incorporates base-policy initialization: We first rollout the base policy for random steps, then let the learned residual RL policy to take over, resulting in demonstration trajectories $\tau_{demo} = \{ (s_1, a_{b,1}),\dots, (s_{t-1}, a_{b,t-1})\} \cup \{(s_t, a_{b,t}+\bar{a}_t), \dots \}$ that contain the behavior of the expert recovering from a potential suboptimal region. We refer to this procedure as \textbf{base policy probing}. Accordingly, we boost the robustness of the RL expert by training the RL expert on an initial state distribution $s_0 \sim p_0^{\pi_b}$ given by random steps of base policy probing. The probing step only serves as state initialization and will not be added to the replay buffer. The details of \method are summarized in~\Cref{alg: pipeline}.

%% file: sections/5_experiments_arxiv.tex
\begin{figure}
    \label{fig: data diversity}
    \centering
    \includegraphics[width=0.95\linewidth]{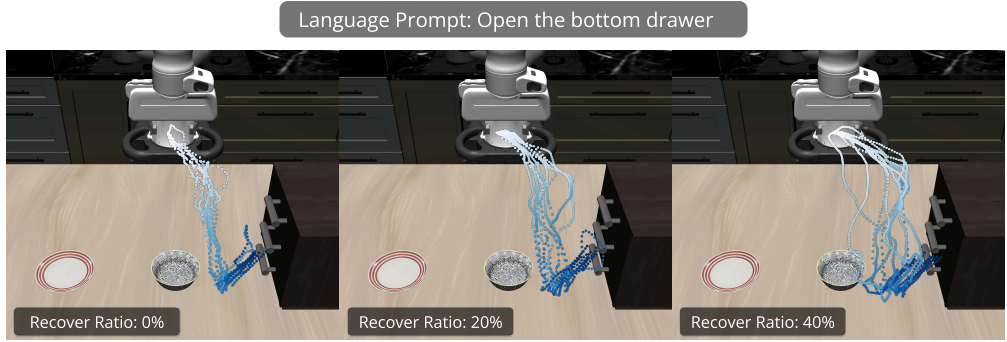}
    \caption{\textbf{Visualization of Data diversity.} We visualize \method data with different base policy initialization probing horizons. Increasing probing horizon yields longer episodes and greater diversity among successful trials. This broader data support leads to improved fine-tuning performance, which eventually saturates.} 
    \label{fig: visual}
\end{figure}

\section{Experiments}

In this section, we systematically evaluate \method. We first demonstrate the efficiency of \method-RL in solving sparse-reward manipulation tasks, which serves as the cornerstone of our pipeline. Then we focus on study 1) How does the probing mechanism of \method benefit VLA SFT; 2) How does \method data compare with other sources of demonstrations (\eg human data, RL expert rollout, VLA base policy rollout). Finally, we investigate the key factors of our pipeline and how they contribute to improving the performance of VLA.

We consider simulation as a proxy to real-world performance, and evaluate methods across two widely adopted simulation benchmarks, including \textbf{LIBERO}~\citep{liu2023libero}, \textbf{SimplerEnv}~\citep{li2024simplerenv}. LIBERO is a lifelong learning benchmark focused on language-guided manipulation tasks. It comprises 130 language-conditioned manipulation tasks grouped into four suites that stress object distribution, spatial arrangement, task goals, and their mixture. SimplerEnv is a robotics manipulation benchmark that aims for high sim-to-real correlation.

In the following sections, we analyze different data sources: \method data $\mathcal{D}^\text{PLD}$, Human data $\mathcal{D}^\text{Human}$, RL expert data (RL expert rollout w/o base policy probing) $\mathcal{D}^\text{RL}$, and base-policy rollout data (Selective successful rollouts, also referred to as "self-bootstrap data'') $\mathcal{D}^\text{Base Policy}$. 
Unless stated otherwise, all methods use identical data volume, training budgets, augmentation, and hyperparameters across architectures; the default base policy we used is $\pi_0$~\citep{black2024pi0}.

\subsection{Effectiveness and Efficiency of Learning RL Specialist}
\label{subsec: RL exp}
\begin{figure}
    \centering
    \includegraphics[width=0.95\linewidth]{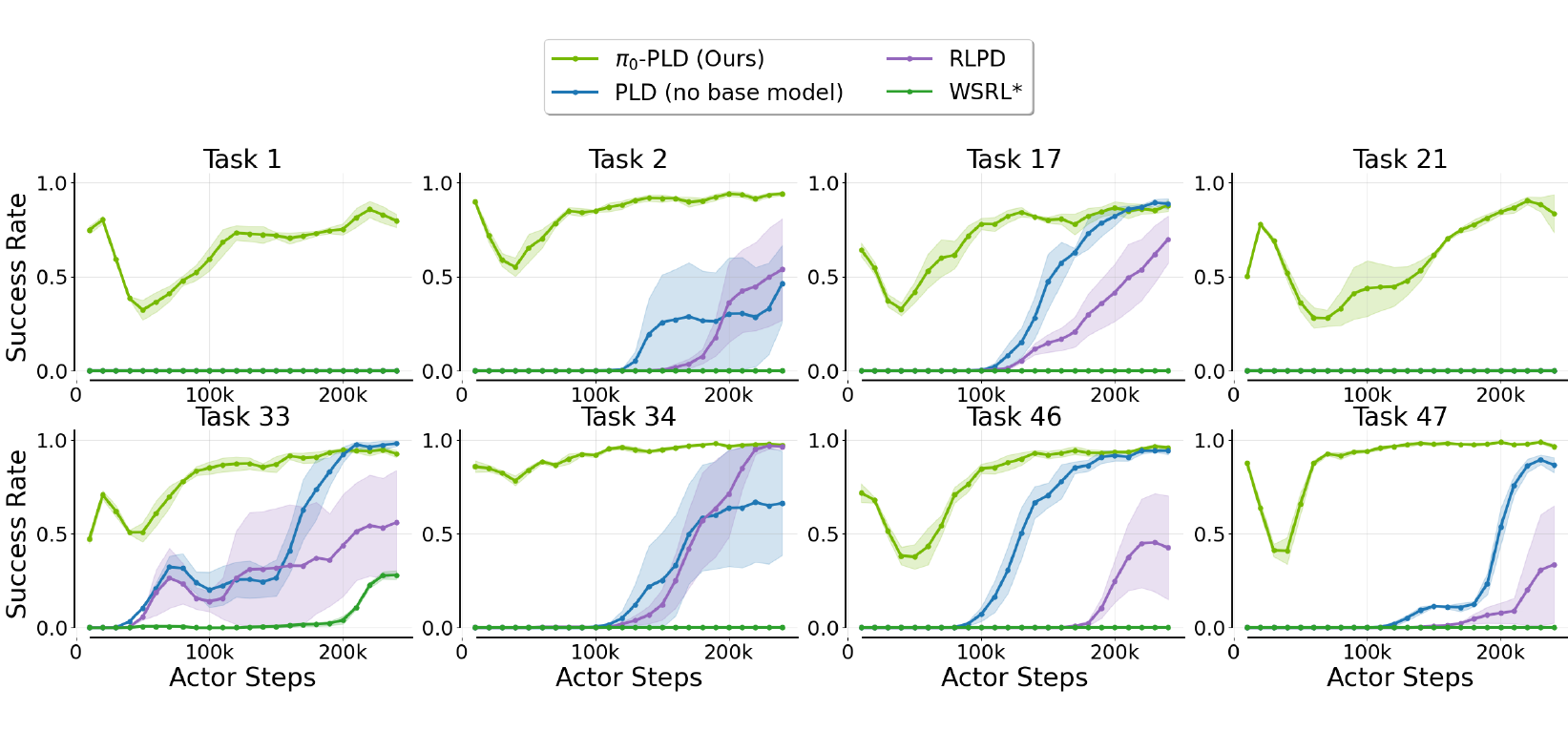}
    \caption{\textbf{Benchmarking Sample-Efficient RL Performance.} We compare \method with RL baseline algorithms that either leverage policy prior or data prior. We report mean rollout performance (Average return calculated within a sliding window of 100 episodes) and 95\% CIs for 3 seeds across 8 manipulation tasks selected from LIBERO-90.}
    \label{fig: RL performance}
    \vspace{-6mm}
\end{figure}

In this section, we seek answers to the following questions: Does \method benefit from both policy guidance and hybrid online learning? We compare state-of-the-art methods that leverage policy priors and data priors: \textbf{WSRL}~\citep{WSRL} (offline initialization only);
\textbf{RLPD}~\citep{RLPD} (No base policy guidance). For the pre-training stage, we collect a dataset of 50 trajectories per task, containing only the successful trials of the same base policy ($\pi_0$), and using Cal-QL~\citep{cal-QL} as the default pre-training algorithm. Subsequently, we retain these data for methods with online hybrid data replay. We plot the training curve of \(\,250\text{k}\,\) steps of online interaction, showing mean rollout performance and 95\% CIs (confidence level) across 3 seeds in~\Cref{fig: RL performance}.

\method outperforms baseline methods by a large margin across 8 tasks on LIBERO-90, indicating that \method effectively exploits the VLA policy prior and yields pronounced sample efficiency at low interaction budgets. In terms of asymptotic performance, \method can achieve \emph{over 95\% performance} on every task that we report to fine-tune performance (over \textbf{120} manipulation tasks). 
Notably, we observe an initial performance drop for \method. This phenomenon implies the initial phase of exploration, where the residual policy starts to diverge from the base policy and visits potentially suboptimal states. Ablation study on \method-RL's design choice can be found in the~\Cref{sec: RL design}.

\begin{table}[htp]
\centering
\caption{Performance on LIBERO benchmark of VLA models fine-tuned on \method data.}
\label{tab: libero}
\resizebox{0.95\linewidth}{!}{%
\setlength{\tabcolsep}{8pt}
\begin{tabular}{lcccccccc}
\toprule
 & \multicolumn{4}{c}{$\pi_0$} & \multicolumn{4}{c}{OpenVLA} \\
\cmidrule(lr){2-5} \cmidrule(lr){6-9}
\textbf{Model} & \textbf{Spatial} & \textbf{Object} & \textbf{Goal} & \textbf{Avg} 
& \textbf{Spatial} & \textbf{Object} & \textbf{Goal} & \textbf{Avg} \\
\midrule
Baseline (SFT/OFT) & 95.2 & 97.6 & 87.4 & 93.4 
& 92.9 & 99.1 & 83.25 & 91.8 \\
\textit{w/ PLD} & \textbf{97.7} & \textbf{98.5} & \textbf{95.3} & \textbf{97.2} 
& \textbf{99.5} & \textbf{99.1} & \textbf{98.9} & \textbf{99.2} \\
$\Delta$ & \textcolor{nvidiaGreen}{+2.5} & \textcolor{nvidiaGreen}{+0.9} & \textcolor{nvidiaGreen}{+7.9} & \textcolor{nvidiaGreen}{+3.8} 
& \textcolor{nvidiaGreen}{+6.6} & \textcolor{nvidiaGreen}{+0.0} & \textcolor{nvidiaGreen}{+15.7} & \textcolor{nvidiaGreen}{+7.4} \\
\bottomrule
\end{tabular}
}
\end{table}

\begin{table}[h]
\centering
\caption{Evaluate \method on SimplerEnv}
\label{tab: main simplerenv}
\resizebox{\linewidth}{!}{%
\begin{tabular}{lccccc}
\toprule
\textbf{Model} & \textbf{WidowX Pick Eggplant} & \textbf{WidowX Pick Carrot} & \textbf{Google Open Drawer} & \textbf{Google Coke Can} & \textbf{Avg} \\
\midrule
Octo-SFT & 65.5 & 43.3 & 92.5 & 85.7 & 71.8 \\
\textit{w/ ours} & 97.8 & 93.9 & 99.3 & 95.5 & 96.6 \\
$\Delta$ & \textcolor{nvidiaGreen}{+32.3} & \textcolor{nvidiaGreen}{+50.6} & \textcolor{nvidiaGreen}{+6.8} & \textcolor{nvidiaGreen}{+9.8} & \textcolor{nvidiaGreen}{+24.9} \\
\bottomrule
\end{tabular}}
\end{table}

\subsection{In-distribution performance}

\begin{wrapfigure}{r}{0.4\columnwidth}
    \centering
\centerline{\includegraphics[width=0.9\linewidth]{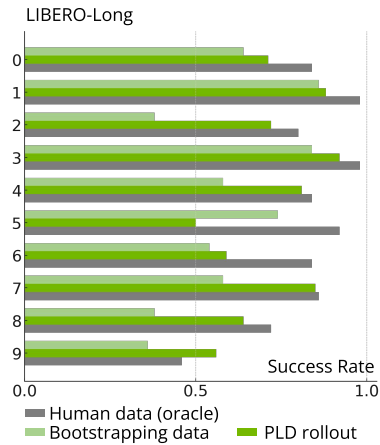}}
    \caption{\textbf{Short-to-long generalization.} $\pi_0$ fine-tuned on LIBERO-90 and one-shot evaluated on LIBERO-10 long horizon tasks.}
    \label{fig: short-to-long}
\end{wrapfigure}

In this section, we investigate how effectively the proposed pipeline enhances the performance of the VLA. We evaluate in-distribution fine-tuning on the LIBERO benchmark using three subsets, each consisting of 10 language-conditioned tasks: \emph{LIBERO-Object}, \emph{LIBERO-Spatial}, and \emph{LIBERO-Goal}. We additionally report results on a custom suite that consists of 4 tasks from SimplerEnv. To demonstrate architecture-agnosticism, we instantiate the base VLA with (i) \textbf{OpenVLA} (autoregressive action tokens)~\citep{kim2024openvla} and (ii) \textbf{$\pi_{0}$} (flow-matching action head)~\citep{black2024pi0}. Since VLA models are mainly trained on real-world datasets that cannot work out of the box on simulation benchmarks, we leverage their official checkpoints for model fine-tuning on each benchmark as the baseline. At test time, each policy is evaluated on \emph{50 episodes per task}, and we report the mean success rate per suite and average over the benchmark. \Cref{tab: libero} and \Cref{tab: main simplerenv} list the performance gain achieved by further applying our method. Across all suites and both architectures, \method data yields consistent absolute gains over human-only SFT while requiring \emph{no additional human demonstrations}. We observe that larger \method datasets monotonically improve in-distribution success and that the distilled generalist notably surpasses the average specialist, indicating effective transfer of task-specific competence into the base VLA.

\begin{figure}
    \centering
    \includegraphics[width=0.95\linewidth]{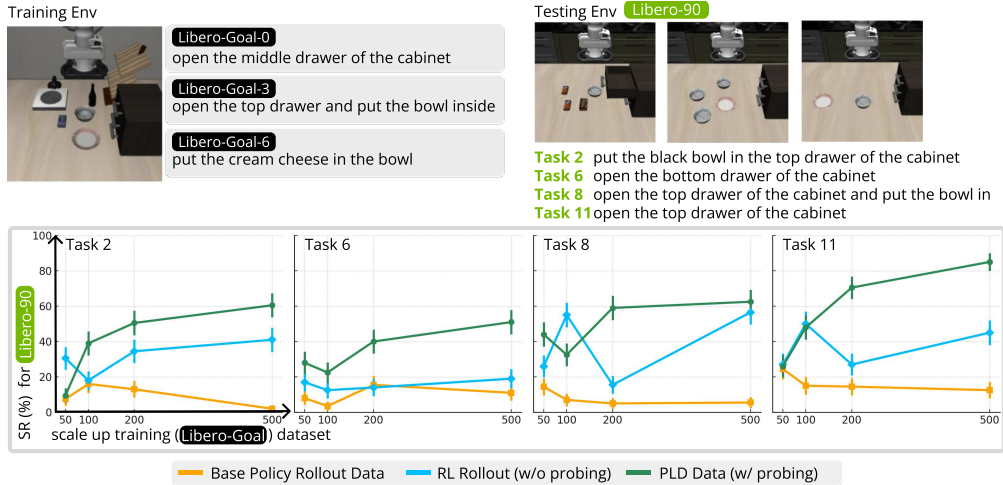}
    \caption{\textbf{Few-shot generalization.} Scaling in-distribution (LIBERO-goal) \method yields better few-shot performance on new tasks (LIBERO-90).}
    \label{fig: three2more}
\end{figure}

\subsection{Generalization}

\paragraph{Generalization to Unseen tasks}
To study the synergetic effect of \method data, we examine whether \method data improves \emph{zero-shot} performance on unseen tasks in the LIBERO benchmark~\citep{liu2023libero}. Concretely, we fine-tune $\pi_0$ via SFT using data drawn from the disjoint \emph{coverage subsets} of LIBERO-90 in proportions \(\{0.1,0.3,0.6,0.8,1.0\}\); for each coverage level, we randomly sample tasks to form a new subset in distribution and then evaluate all tasks in the suite. We sampled 4 subsets for each coverage level to provide a more unified result. We consider three different data sources: (i) Ours $\mathcal{D}^{\textit{PLD}}$, (ii) human expert data $\mathcal{D}^\text{Human}$, and (iii) \emph{self-bootstrapping} data $\mathcal{D}^\text{$\pi_0$ rollout}$ (equal to 0-1 REINFORCE). We visualize the result in~\Cref{fig:seentask-to-unseentask}. Across coverage levels, \(\pi_{0}\) fine-tuned on $\mathcal{D}^{\texttt{PLD}}$ attains the strongest in-distribution performance and maintains robust zero-shot transfer to unseen tasks; human data-only SFT achieves approximately a similar level of zero-shot generalization at the same training budget but lags on in-distribution tasks; $\pi_0$ self-bootstrapping rollout data underperforms in-distribution and fails to generalize to out-of-distribution tasks. 

\paragraph{Generalization to Out-of-domain}
We study \emph{few-shot generalization} for tasks with different goals, layouts, and backgrounds. We first collect \method data of varying scales set on \emph{source} tasks (LIBERO-Goal) and evaluate the fine-tuning performance on \emph{target} tasks (LIBERO-90). Specifically, the VLA is also fine-tuned on a small number of oracle demos of target tasks. To analyze transfer by skill family, we select tasks from LIBERO-goal and LIBERO-90 that have high semantic correlation to form a set of source/target tasks. We scaled the size of \(|\mathcal{D}^{\text{PLD}}|\) from 50 to 500 trajectories and compared against $\mathcal{D}^{RL}$ and $\mathcal{D}^{BS}$ under the same data and training budget. As shown in~\Cref{fig: three2more}, we observe monotonic improvements in SFT performance as the data scales from 50 to 500 trajectories.

\paragraph{Generalize to Long-horizon}
We assess skill composition ability on LIBERO-100 by fine-tuning the base VLA on LIBERO-90 (source) and evaluating one-shot (give one human demo each) on the held-out LIBERO-10 long-horizon tasks (target). To construct \method\ data, we first train residual RL specialists independently on each LIBERO-90 task, then aggregate their successful rollouts. As shown in~\Cref{fig: short-to-long}, the fine-tuning of the data \method exceeds the tuning of the data from the baseline policy roll-out (self-bootstrapped), but still falls short of the performance achieved with demonstrations by human experts.

\begin{figure}[htp]
\begin{center}    
    \includegraphics[width=1.0\linewidth, keepaspectratio]{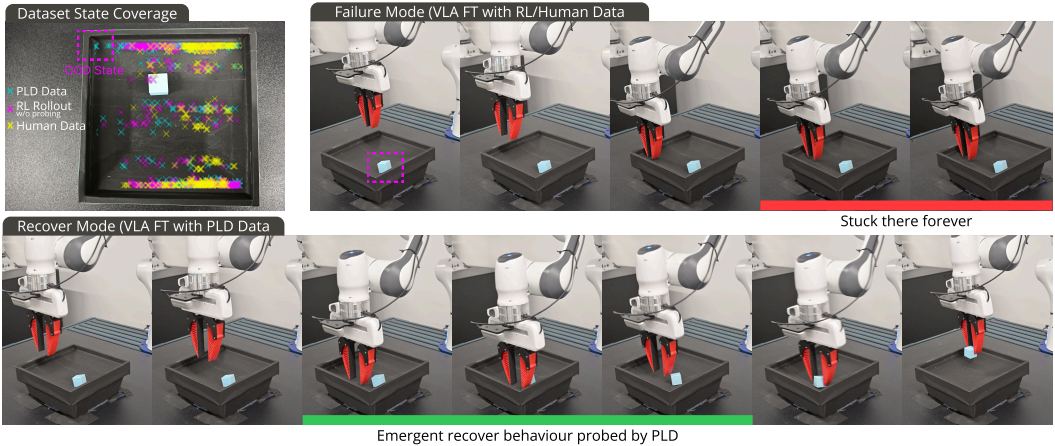}
    \caption{\textbf{Visualization of failure mode and recovery behavior in the real-world.}}
    \label{fig:real-world-failure-mode}
\end{center}
\end{figure}

\subsection{Real-world Performance}
\label{subsec: real}

We evaluate our approach on a 7-DoF Franka Emika Panda arm in the real world, considering two sets of canonical manipulation tasks: \emph{pick-and-place} and \emph{peg insertion}, illustrated in~\Cref{fig: real-world-setup}. Unlike prior works~\citep{luo2025hilserl, WSRL}, we do not restrict task randomization, making real-world reinforcement learning particularly challenging. A more detailed experimental setup is provided in~\Cref{sec:real-world-setup}.

\paragraph{Data collection and policy training.}
We first collected 200 teleoperated trajectories to perform supervised fine-tuning (SFT) of the base policy $\pi_0$. Using this initialization, we trained $\pi_0$-\method and $\pi_0$-RLPD without human interventions. Both policies reached 100\% success on the two tasks within 2 hours of training. We then leveraged the learned expert policies to autonomously collect 200 successful demonstrations each, forming datasets $\mathcal{D}^{\text{PLD}}$ and $\mathcal{D}^{\text{RLPD}}$, which were subsequently used to further SFT $\pi_0$, yielding $+\mathcal{D}^\text{PLD}$, $+\mathcal{D}^\text{Human}$, and $+\mathcal{D}^\text{RLPD}$.

\paragraph{Performance and failure modes.}  
Across 30 randomized trials per task, all methods achieved perfect success on peg insertion (30/30), demonstrating robust reactive skills. In cube pick-up, however, $+\mathcal{D}^\text{RLPD}$ and $+\mathcal{D}^\text{Human}$ succeeded in only 16/30 and 10/30 trials, respectively, while $+\mathcal{D}^\text{PLD}$ maintained 30/30. \Cref{fig:real-world-failure-mode} illustrates a typical failure: policies trained on $\mathcal{D}^{\text{RLPD}}$ or $\mathcal{D}^{\text{Human}}$ often pushed the cube into the upper-left corner, where the gripper became stuck. By contrast, $+\mathcal{D}^\text{PLD}$ was reliably recovered by repositioning the cube before grasping. Distribution analysis confirms that neither human demonstrations nor RL rollouts visited such corner states, whereas \method\ explicitly probed the base policy and generated diverse trajectories that captured these cases. This explains its robustness and highlights its potential as a self-improving data flywheel. We also consider a more challenging setting of the randomized evaluation environment, demonstrating the generalizability of \method even in the real world. Details can be found in~\Cref{sec: generalization}.

\paragraph{Robustness for long-horizon tasks.}
To evaluate the robustness of \method in executing long-horizon and dexterous manipulation tasks, we set up two 6-DoF YAM robot arms developed by~\cite{i2rt_yam_arm}. We consider an industrial insertion task—specifically, inserting a micro graphics card into a motherboard. To enable fully autonomous operation without human intervention or resetting, we decompose the task into four stages: Stage 1: Pick up the GPU from the table and insert it into slot 1. Stage 2: Move the GPU from slot 1 to slot 3. Stage 3: Firmly insert the GPU into slot 3. Stage 4: Unplug the GPU from slot 3 and place it back on the table. A reward classifier is trained to govern the state machine that coordinates these stages. After at most 8 hours of training for each subtask and distilling the learned skills into a single BC base policy, the system can continuously perform the full task loop without human assistance for at least 1 hour. As shown in the video, although the one-shot success rate for each stage is not 100\%, the system is capable of recovering from failures, keeping the data flywheel running autonomously.


\begin{figure}
    \centering
    \includegraphics[width=0.9\linewidth]{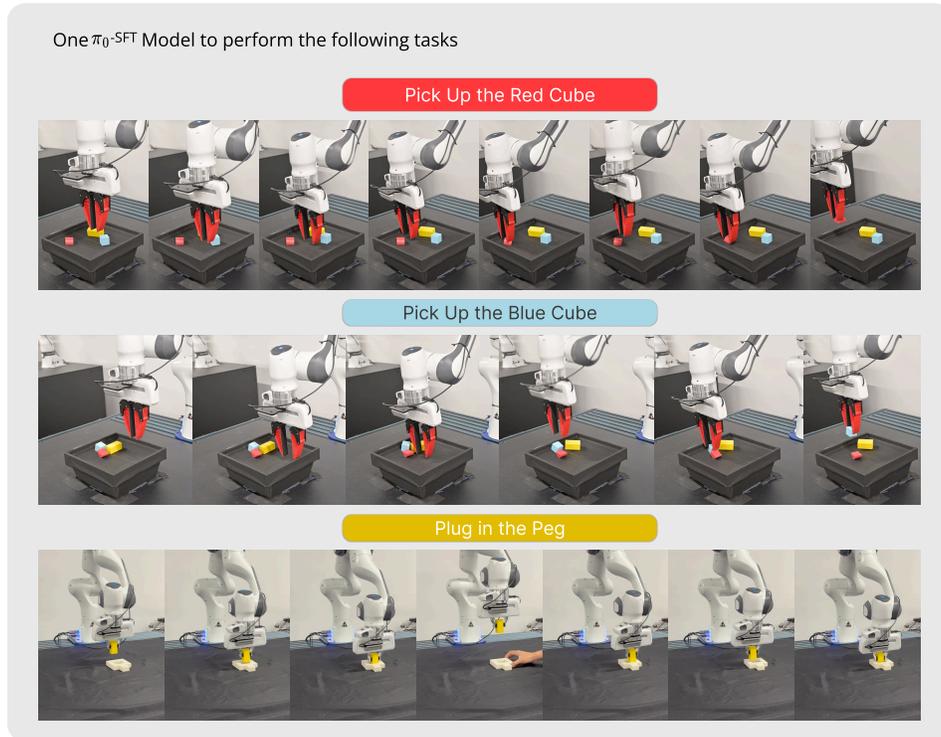}
    \caption{\textbf{Real-world Generalization Performance.} We evaluate one model's multi-task performance on three language-conditioned manipulation tasks, including pick-and-place and peg insertion.}
    \label{fig:real-world-generalization}
\end{figure}

\begin{figure}[htp]
    \centering
    \includegraphics[width=0.95\linewidth]{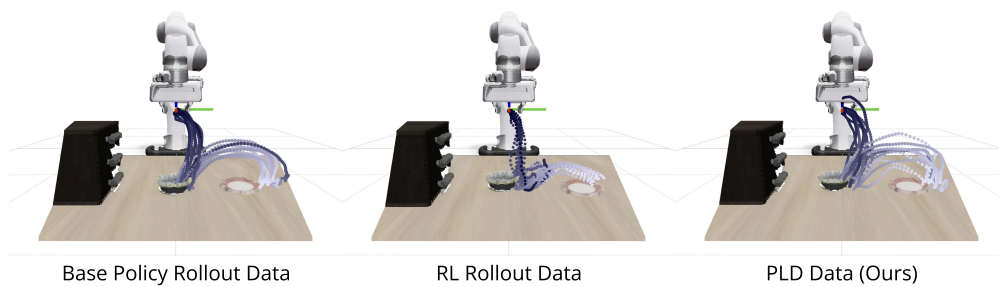}
    \caption{\textbf{Visualization of different data sources.} We plot 20 trajectories for each method (task prompt: ``pick up the black bowl and place it on the plate''). RL expert data is of high quality but lacks diversity and diverges far from base policy behavior, while \method data aligns better with the base policy and contains diverse recovery behavior. (More interactive visualizations at~\url{\pagelink}).}
    \label{fig:vis-compare-trajectory}
\end{figure}

\subsection{How does \method work?}
\label{subsec: explanation}
We take a deeper look at the underlying reason for \method data's bonus in generalization. As shown in~\Cref{fig:vis-compare-trajectory}, we plot 50 trajectories for each method (task description: ``open middle drawer of the middle cabinet''). RL expert provides optimal and concentrated solutions to the task, but lacks diversity and diverges far from the behavior of the base policy, while \method data are clustered near the trials of the base policy and contain various recovery behaviors. Based on empirical observation, we hypothesize that due to the base policy probing, \method data provides a solution that is biased towards the base policy, thus fine-tuning forgets less of the base model's generalizability. This resembles observations in LLM fine-tune~\citep{shenfeld2025razor}, where the KL-divergence can serve as an indicator of forgetting. Meanwhile, large data coverage also benefits robustness in sequential decision making~\citep{kelly2019hg}.

\begin{wrapfigure}{r}{0.4\columnwidth}
    \vspace{-4mm}
    \centering
    \centerline{\includegraphics[width=0.95\linewidth]{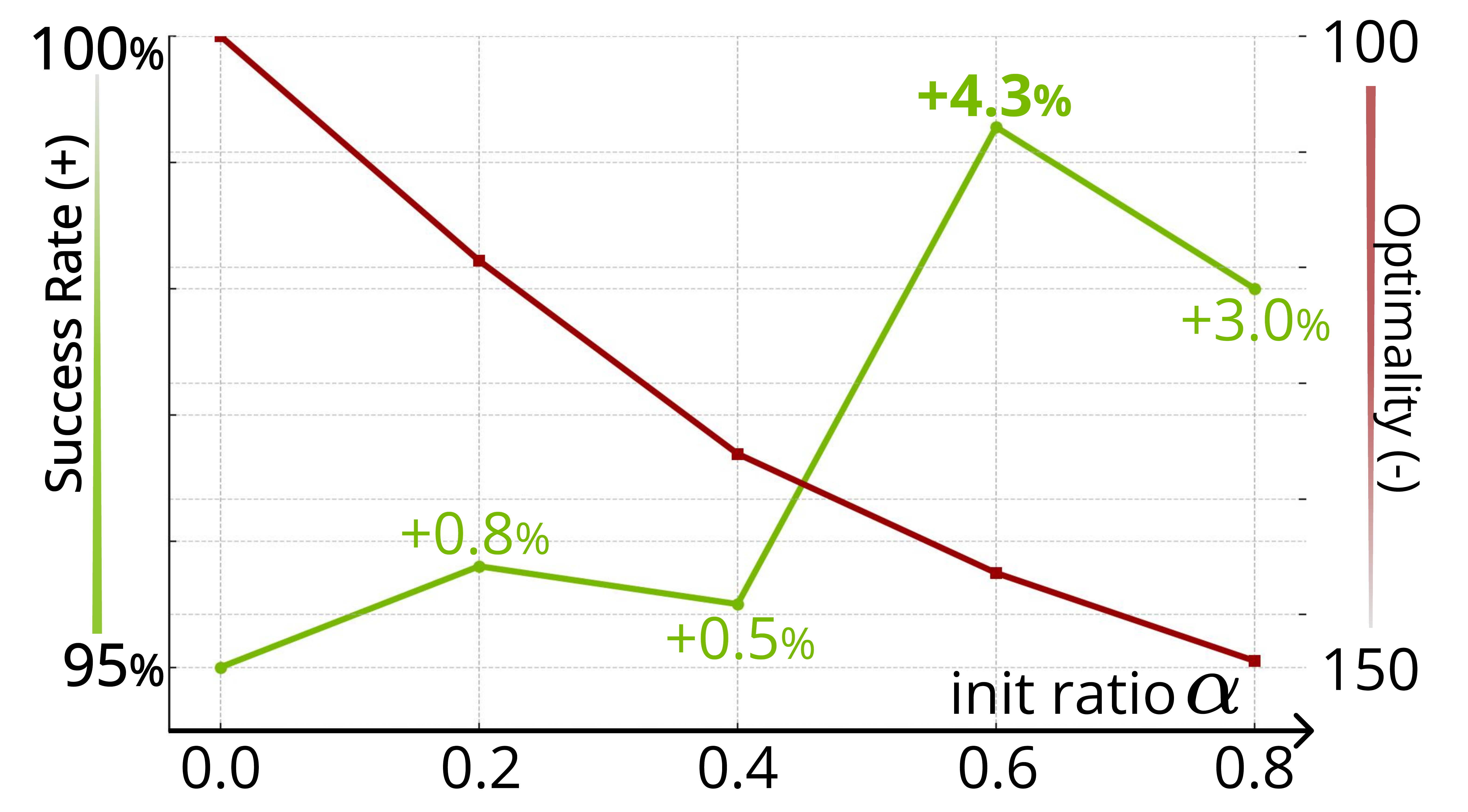}}
    \caption{\textbf{Ablation of Probing Horizon.} Performance plateau at $\alpha=0.6$.}
    \label{fig:random-probe-ratio}

\end{wrapfigure}

In addition, we study sensitivity to the initialization horizon. We choose task 0-9 from LIBERO-90, change the steps we used to initialize the random sample, initiating steps $T_\text{base} \sim [0, \alpha T]$ to rollout the base policy. $\alpha \in [0.0, 0.2, 0.4, 0.6, 0.8]$. As $\alpha$ increases, the average episode length of successful trajectories increases, indicating a detour required to correct the suboptimal behavior of the base policy. As demonstrated in~\Cref{fig:random-probe-ratio}, performance plateaus at $\alpha=0.6$ and drops as $\alpha$ increases further. This is consistent with our analysis that SFT benefits from the data diversity.

%% file: sections/2_related_works.tex
\section{Related Work}
\label{sec: related works}
\subsection{Robotics Foundation Models}

Following the success of large language models and vision language models~\citep{brown2020gpt3, touvron2023llama2, chen2022pali}, recent works on robotics foundation models turned to a similar transformer-based architecture with aggressive data scaling. This inspired earlier works in VLAs such as RT-1, RT-2, and OpenVLA, etc. ~\citep{brohan2023rt1, brohan2023rt2, kim2024openvla, xue2025leverb}. Meanwhile, diffusion-based action generation, explored in \cite{chi2024diffusionpolicy}, takes motivation from generative modeling techniques~\citep{ho2020denoisingdiffusionprobabilisticmodels}, demonstrating smooth and accurate action generation. This has led to more recent VLA architectures to-date, such as Octo~\citep{team2024octo}, OpenVLA-OFT~\citep{openvla-oft}, GR00T~\citep{bjorck2025gr00t}, and the $\pi$-series of models \citep{black2024pi0, intelligence2025pi05, pertsch2025fast}. The VLA training procedure is typically analogous to VLM training. First, model weights are initialized from the respective VLM backbones~\citep{kim2024openvla, black2024pi0}. Then, the model is supervised with next-token-prediction tasks on diverse pretraining datasets, spanning across multi-modal web data~\citep{intelligence2025pi05} such as COCO~\citep{chen2015microsoftcococaptionsdata} and VQAv2~\citep{goyal2017making}, and robotics-specific, cross-embodiment data~\citep{khazatsky2025droid,o2024open}. Finally, supervised fine-tuning is conducted on a small set of high-quality teleoperation data collected from the target robot deployment platform performing the target tasks. 

\subsection{Sample-efficient RL with Data and Policy priors} 
Sample and exploration efficiency have been a long-standing problem in RL, especially in sparse-reward settings. Recent works have explored leveraging offline data to improve sample efficiency. Offline-to-online transfer~\citep{vecerik2017leveraging, nair2020awac, IQL, cal-QL, WSRL, li2025Qchunking} considers a two-stage pipeline that first initializes policy or critic using pessimism or constrained objective in offline RL~\cite{levine2020offlineRLtutorial} and follows with an online fine-tuning phase to have new data collected and alleviate distributional shift; Hybrid RL~\citep{song2022hybrid, song2024hybrid, RLPD} considers online RL with access to an offline dataset. Given expert demonstration, one can either continuously replay this data to ensure high-value state visitation~\cite{RLPD} or to guide exploration~\cite{dong2025DGN}. Data prior can also guide reset-free real-world learning~\citep{walke2023ariel, sharma2023self}.
Another line of work assumes access to policy prior, such as a pre-trained generalist. \cite{ye2023FAC, chen2025conrft, julg2025RPD} leverage foundation policy to guide RL through an auxiliary behavior regularization objective. Action editing is another efficient way to improve upon the policy prior. ResiP~\citep{ankile2025imitation} considers learning a residual policy through PPO~\citep{PPO}, while EXPO~\citep{dong2025expo} considers an off-policy solution and co-trains the base policy during the process. Our work leverages a suboptimal base policy to achieve a non-zero success rate for warm-starting exploration, but does not require access to oracle demos or a human expert for further intervention.

\subsection{VLA post-training}
The prevailing large-scale recipe for VLA post-training is to \emph{pretrain} on diverse, heterogeneous robot data and then \emph{fine-tune} on task-specific demonstrations~\citep{zhou2024autonomous,black2024pi0}. For example, \citet{black2024pi0} performs supervised post-training on a carefully curated task-targeted corpus, with per-task coverage ranging from a few to over 100 hours of teleoperation. Because such post-training data are expensive to acquire, the authors note that most diversity must come from the pretraining mixture—underscoring a key limitation of pure SFT: data scarcity and limited coverage at adaptation time. To enable self-improvement, prior work has explored scaling high-quality data via \emph{online RL specialists}~\citep{RLPD}. However, these pipelines often require substantial human intervention and collect data in a large way \emph{agnostic} to the generalist's behavior, restricting scalability. Other lines investigate \emph{on-policy} RL for post-training~\citep{lu2025vla-rl,tan2025ript-vla}, or optimize \emph{single-task} fine-tuning at the expense of generalization~\citep{chen2025conrft}. Our work jointly targets these limitations by seeking a post-training pipeline that reduces human effort, aligns data collection with the generalist’s state distribution, and remains sufficiently sample efficient for real-world systems.

%% file: sections/6_conclusions.tex
\section{Conclusions}
We presented \method, a three-stage post-training pipeline that enables VLA models to improve autonomously without relying on additional oracle human demonstrations. \method couples a frozen VLA generalist with lightweight \emph{residual} RL specialists to warm-start exploration and distills curated successes back into the base model with standard SFT. Across large-scale simulation experiments and real-world deployment, \method improves without additional human demonstration, achieving near-saturated $\sim$99\% success on LIBERO, $>$50\% gains in SimplerEnv, and robust real-world performance. Ablations identify \emph{residual policy probing} and \emph{distribution-aware replay} as key to stability, sample efficiency, and generalization. We consider \method{} as a practical step toward autonomous, scalable post-training and a foundation for future work on multi-embodiment transfer, continual on-robot learning, and safety-constrained data collection.

%% file: sections/8_acknowledgement.tex
\section{Acknowledgments}
We are grateful to Jason Liu, Tony Tao, Colin Li, Max Fu, Yuhui Chen, Ajay Mandlekar, You Liang Tan, Dennis Da, Haoyu Xiong, Stephanie Chen, Charles Xu, and Guanzhi Wang for their insightful discussions and technical support. We also thank Tri Cao, Jeremy Chimienti, and Lion Park for their assistance with data collection and mechanical setup. Finally, we thank the GEAR Team and LeCAR Lab for their continuous support.

%% file: sections/7_appendix_arxiv.tex
\section{Algorithm}
\begin{algorithm}[] 
   \caption{\texttt{PLD} with base-policy initialization}
   \label{alg: pipeline}
\begin{algorithmic}
   \REQUIRE $\pi_b$, $\pi_\delta$, $Q_\phi$, $Q_{\phi '}$, $\alpha$, $\gamma$, $\mathcal{B}_\text{offline}$, $\mathcal{B}_\text{online}$
    \STATE \textbf{\# Initialization}
    \STATE Collect $n$ successful trials of $\pi_b$: $\mathcal{D}_{offline} = \{\tau_1, \tau_2, \dots\, \tau_n\}$
    \STATE Initialize online buffer $\mathcal{D}_{online} = \varnothing$
    \STATE Initialize the critic network $Q_\phi, Q_{\phi'}$ with Cal-QL on $\mathcal{D}_{offline}$
    \STATE Randomly initialize delta policy network $\pi_\delta$
    \STATE \textbf{\# RL training}
    \STATE Freeze $\pi_b$, denote $\bar\pi(\cdot|s) = \pi_b(\cdot|s)\pi_\delta(\cdot | s, a_b)$
    \FOR{each RL step}
        \IF{collect data}
            \IF{Warm up step}
                \STATE base model rollout $ a \sim \pi_{base}(\cdot | s)$
            \ELSE
                \STATE sample action $\bar{a} \sim \bar\pi(\cdot | s)$
            \ENDIF
            \STATE Environment step: $r, s', done = env.step(\bar{a})$
            \STATE Add $(s, a, \mu, r, s')$ to buffer $\mathcal{D}_{online}$.
        \ENDIF
        \STATE Equally sample data from online and offline buffer: $b \sim \mathcal{D}_{online} \cup \mathcal{D}_{offline}$
        \STATE Calculate TD target by bootstrapping $\bar{\pi}$
        \STATE Update $Q_\phi$ by~\Cref{eqn: Q def}
        \STATE Update $\pi_{\delta}$ by maximizing the SAC target
        \STATE \textbf{Polyak update} $\phi'=\rho\phi' + (1-\rho)\phi$
    \ENDFOR
    \STATE \textbf{\#Base policy SFT}
    \STATE For each task, we collect hybrid behavior dataset $\mathcal{D}_{SFT}$: \[
                \pi(s_t) =
                \begin{cases} 
                a_{base}, & t < T_{base} \\ 
                a_{base} + a_{\delta} , & t \geq T_{base}
                \end{cases}
                \]
    \FOR{each SFT step}
        \STATE update $\pi_b$ by BC objective.
    \ENDFOR
    \STATE Return $\pi_b$
\end{algorithmic}
\end{algorithm}

\begin{table}[h]
\centering
\caption{LIBERO-90 Success rate for $\pi_0$ SFT with different dataset.}
\label{tab:libero-90-seen-to-unseen}
\setlength{\tabcolsep}{4pt} 
\renewcommand{\arraystretch}{1.1}

\begin{minipage}{0.31\linewidth}
\centering
PLD Data\\
\scriptsize
\begin{tabular}{ccc}
\toprule
Ratio & Overall SR & Seen/Unseen \\
\midrule
0.1 & 0.314 & \su{0.941}{0.244} \\
0.3 & 0.470 & \su{0.968}{0.259} \\
0.6 & 0.637 & \su{0.872}{0.283} \\
0.8 & 0.745 & \su{0.864}{0.268} \\
1.0 & 0.871 & \su{0.871}{N/A} \\
\bottomrule
\end{tabular}
\end{minipage}\hspace{0.01\linewidth}
\begin{minipage}{0.31\linewidth}
\centering
Base Policy Rollout Data\\
\scriptsize
\begin{tabular}{ccc}
\toprule
Ratio & Overall SR & Seen/Unseen \\
\midrule
0.1 & 0.103 & \su{0.523}{0.056} \\
0.3 & 0.068 & \su{0.198}{0.015} \\
0.6 & 0.328 & \su{0.506}{0.062} \\
0.8 & 0.344 & \su{0.423}{0.031} \\
1.0 & 0.488 & \su{0.488}{N/A} \\
\bottomrule
\end{tabular}
\end{minipage}\hspace{0.01\linewidth}
\begin{minipage}{0.31\linewidth}
\centering
Human Data\\
\scriptsize
\begin{tabular}{ccc}
\toprule
Ratio & Overall SR & Seen/Unseen \\
\midrule
0.1 & 0.272 & \su{0.796}{0.214} \\
0.3 & 0.419 & \su{0.829}{0.240} \\
0.6 & 0.611 & \su{0.829}{0.286} \\
0.8 & 0.694 & \su{0.803}{0.256} \\
1.0 & 0.815 & \su{0.815}{N/A} \\
\bottomrule
\end{tabular}
\end{minipage}

\end{table}

\input{asset/tables/task_description}

\section{Implementation}

\subsection{RL Baselines}
\label{sec: RL baselines mod}
To ensure an apple-to-apple comparison in~\Cref{subsec: RL exp}, we implement these baselines based on the SERL~\cite{luo2024serl} framework and adapt them to fit in the settings of our study. We provide a detailed explanation of baseline formulation and our implementation.

\paragraph{RLPD}
RLPD~\citep{RLPD} proposed a hybrid RL pipeline that leverages offline data to foster learning in challenging sparse reward settings. During training, it equally draws samples from both the online and offline buffer. It also uses LayerNorm to deal with the Q-value blow-ups common when querying OOD actions under a high update-to-data (UTD) ratio. We refer to the implementation in the SERL software for both simulation and real deployment.

\paragraph{WSRL} 
In the original paper~\citep{WSRL}, WSRL uses Cal-QL~\citep{cal-QL} to pre-train both the action and critic during the offline phase. For the online phase, it discards offline data and warms up the replay buffer with 50k steps of pre-trained policy rollouts.
We did not provide a large, diverse dataset like D4RL~\citep{fu2020d4rl} benchmark. Rather, we use the same procedure as \method to collect successful trajectories from the base model. We implement WSRL under the SERL framework, as the UTD is no longer fixed to 4. This baseline can be considered as an ablation of the residual policy and offline data replay. We use the WSRL baseline as an ablation study of the warm-up online exploration using the base policy, and offline data retention through hybrid data replay.

\paragraph{JSRL}
Jump-start RL~\citep{JSRL} is a meta-algorithm using an existing guide policy to “rolling-in”. The key mechanism is to shape the initial-state distribution for the learner: JSRL repeatedly resets episodes from states that the guide visits (a curriculum from easy/near-goal states to harder/far-from-goal states), making difficult tasks learnable with fewer trials. It leverages the guide policy for data collection, without directly imitating its actions. JSRL is agnostic to the underlying RL backbone. In practice, we choose SAC to learn the exploration policy. Since JSRL only leverages the policy prior (VLA policy in practice) to warm-up exploration during online interaction, we use it as an ablation of the hybrid experience replay mechanism. 

\paragraph{Cal-QL}
Calibrated Q-learning~\citep{cal-QL} addresses the underestimation issue of CQL~\citep{CQL}, thereby significantly improving fine-tuning performance in the offline-to-online setting. It learns a conservative value function that underestimates the value of OOD actions, while ensuring the values are within a reasonable scale. In practice, it under-bounds the conservative Q function by the value of the behavior policy $\mu$ (policy corresponds to the offline dataset $\mathcal{D}$). The modified Q-learning objective is the following:
\[
    \min_{\theta} \alpha \left( \mathbb{E}_{s\sim \mathcal{D}, a \sim \pi}[\max(Q_\theta (s, a), V^\mu(s))] \right) - \frac{1}{2} \mathbb{E}_{s, a \sim \mathcal{D}} \left[ (Q_\theta(s, a) - \mathcal{B}^\pi \bar{Q}(s, a))^2 \right],
\]
where $\bar{Q}$ is the target Q-value function and the second term corresponds to minimizing TD-error~\cite{lillicrap2015continuous}.

\paragraph{Implicit Q-Learning (IQL).}
IQL is an in-sample offline RL method that avoids querying $Q$ on out-of-distribution actions while still improving over the behavior policy~\citep{IQL}. The key step is to fit a \emph{state value} $V_\psi$ by \emph{expectile regression} over the actions of the dataset and then bootstrap $Q_\theta$ toward this value. Let $\delta(s,a)=Q_\theta(s,a)-V_\psi(s)$ and define the expectile loss
$\mathcal{L}_\eta(\delta)=\lvert \eta-\mathbf{1}\{\delta<0\}\rvert\,\delta^2$ with $\eta\in(0.5,1)$.
IQL alternates
\begin{align}
\text{(V)}\quad
&\min_{\psi}\ \mathbb{E}_{(s,a)\sim\mathcal{D}}\big[\mathcal{L}_\eta\!\big(Q_\theta(s,a)-V_\psi(s)\big)\big],\\
\text{(Q)}\quad
&\min_{\theta}\ \mathbb{E}_{(s,a,s')\sim\mathcal{D}}\Big[\big(Q_\theta(s,a)-\big(r+\gamma V_\psi(s')\big)\big)^2\Big],\\
\text{(policy)}\quad
&\max_{\phi}\ \mathbb{E}_{(s,a)\sim\mathcal{D}}\!\Big[\exp\!\big(\tfrac{Q_\theta(s,a)-V_\psi(s)}{\beta}\big)\,\log \pi_\phi(a\mid s)\Big],
\end{align}
which realizes policy improvement without out-of-distribution action queries (the policy step reduces to advantage-weighted regression)~\citep{IQL}.
In our comparison to \emph{Cal-QL}~\citep{cal-QL} as a critic-initialization baseline, we consider a simplified version of IQL that \emph{directly} regresses $Q_\theta$ toward an $n$-step return
$R_t=\sum_{i=t}^{t+n-1}\gamma^{\,i-t}r_i+\gamma^{\,n}V_\psi(s_{t+n})$
using expectile regression:
\[
\min_{\theta}\ \mathbb{E}_{(s_t,a_t)\sim\mathcal{D}}\!\left[\mathcal{L}_\eta\!\big(Q_\theta(s_t,a_t)-R_t\big)\right].
\]
Unless otherwise noted, we set $\eta=0.7$, a value shown to propagate high-value signals effectively in the IQL paper.

\subsection{Design Choices of \method}
\label{sec: RL design}

In this section, we provide a detailed study of design choices that make \method data efficient and achieve high convergence performance. We evaluate all algorithms on the selected 8 LIBERO-90 tasks.

\label{subsec: appendix-real}

\begin{figure}
    \label{fig: ablation reward shaping}
    \centering
    \includegraphics[width=0.85\linewidth]{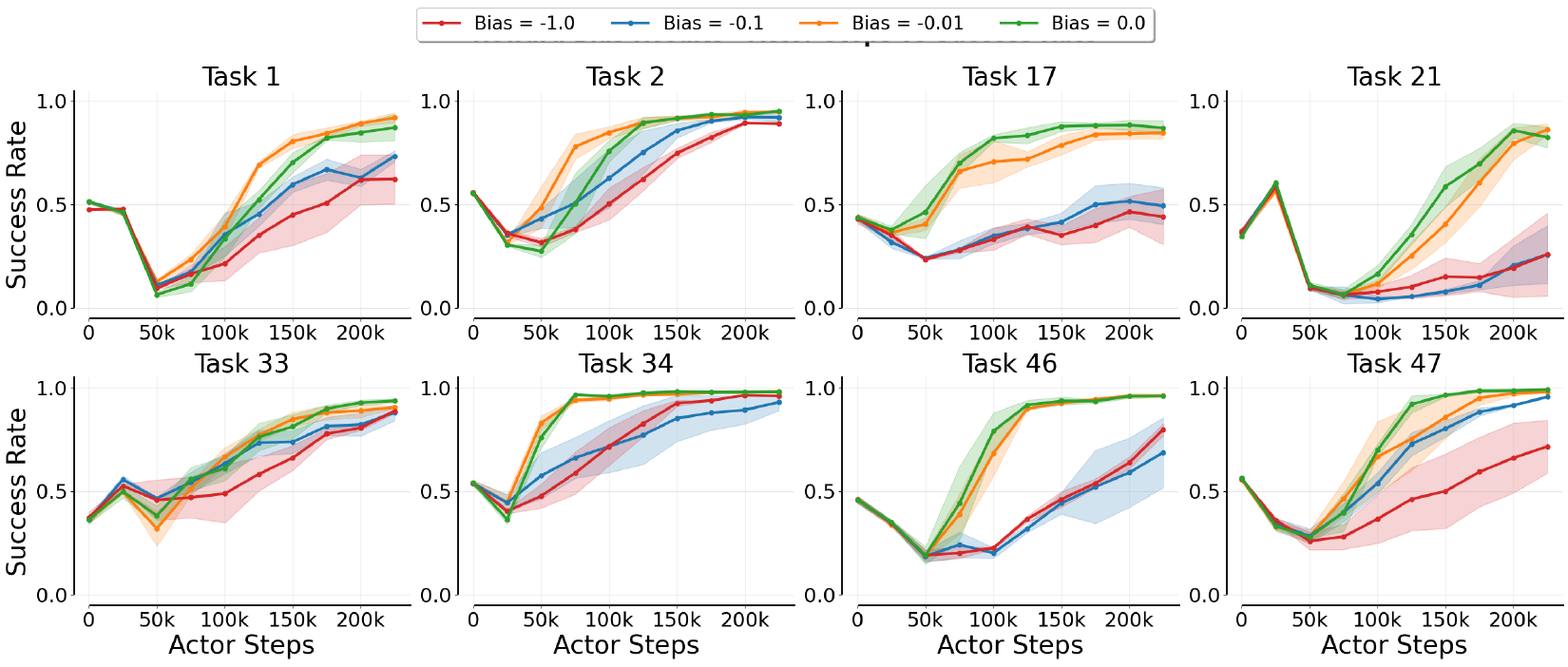}
    \vspace{-20mm}
    \caption{\textbf{Reward bias ablation.} Mean and 95\% CIs of rollout performance across 3 seeds.}
    \label{fig:reward-shaping}
\end{figure}

\paragraph{Reward shaping}
We empirically analyze the impact of naive reward shaping. Specifically, we consider a step-wise \emph{survival cost} as reward bias as in prior works~\cite{luo2024serl}. As shown in~\Cref{fig:reward-shaping}, adding a slight reward bias has little impact, but it could increase convergence speed in 2 out of 8 tasks; However, A large bias could significantly hinder performance. For the major results reported in the main paper, we do not apply reward shaping.

\begin{figure}
    \centering
    \includegraphics[width=0.85\linewidth]{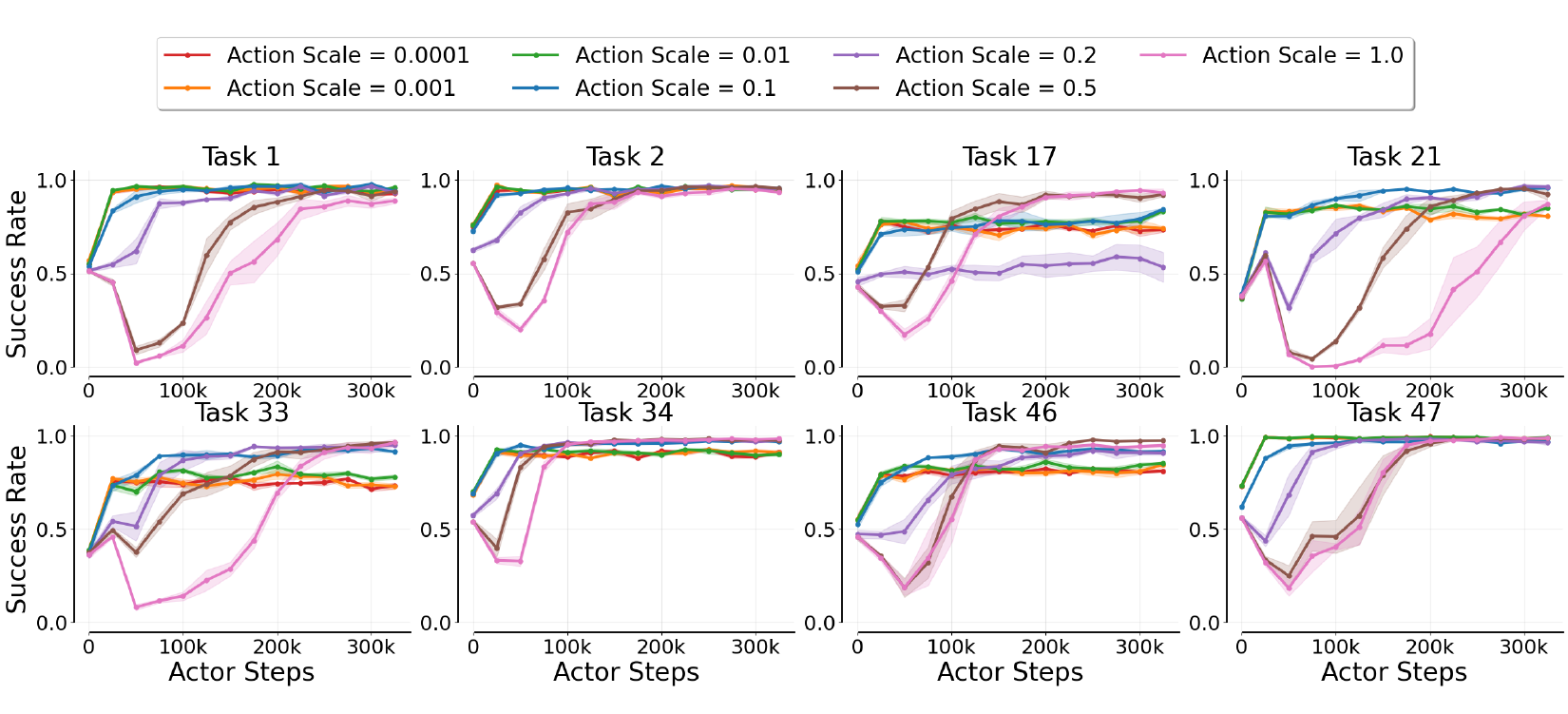}
    \caption{\textbf{Action scale ablation} Mean and 95\% CIs of rollout performance across 3 seeds.}
    \label{fig: ablation action scale}
\end{figure}

\paragraph{Action scale}
One core component of residential policies is the scale of exploration. To avoid unlearning results from diverging too far from the base policy, delta actions are usually scaled down and bounded within a range of \([-\xi, \xi]\)~\citep{ankile2025imitation, dong2025expo}. \Cref{fig: ablation action scale} compares different residual action scales. Setting \(\xi\) too large at the start can degrade early performance: updates deviate excessively from the base policy, inducing unstable exploration, while a small \(\xi\) will lead to insufficient exploration and lower asymptotic performance. We argue that \(\xi\) needs to be carefully tuned to enable exploration while minimizing performance drop. For single-arm manipulation, we suggest \(\xi=0.5\) a good choice for LIBERO and \(\xi=0.1\) for SimplerEnv.

\begin{figure}
    \centering
    \includegraphics[width=0.85\linewidth]{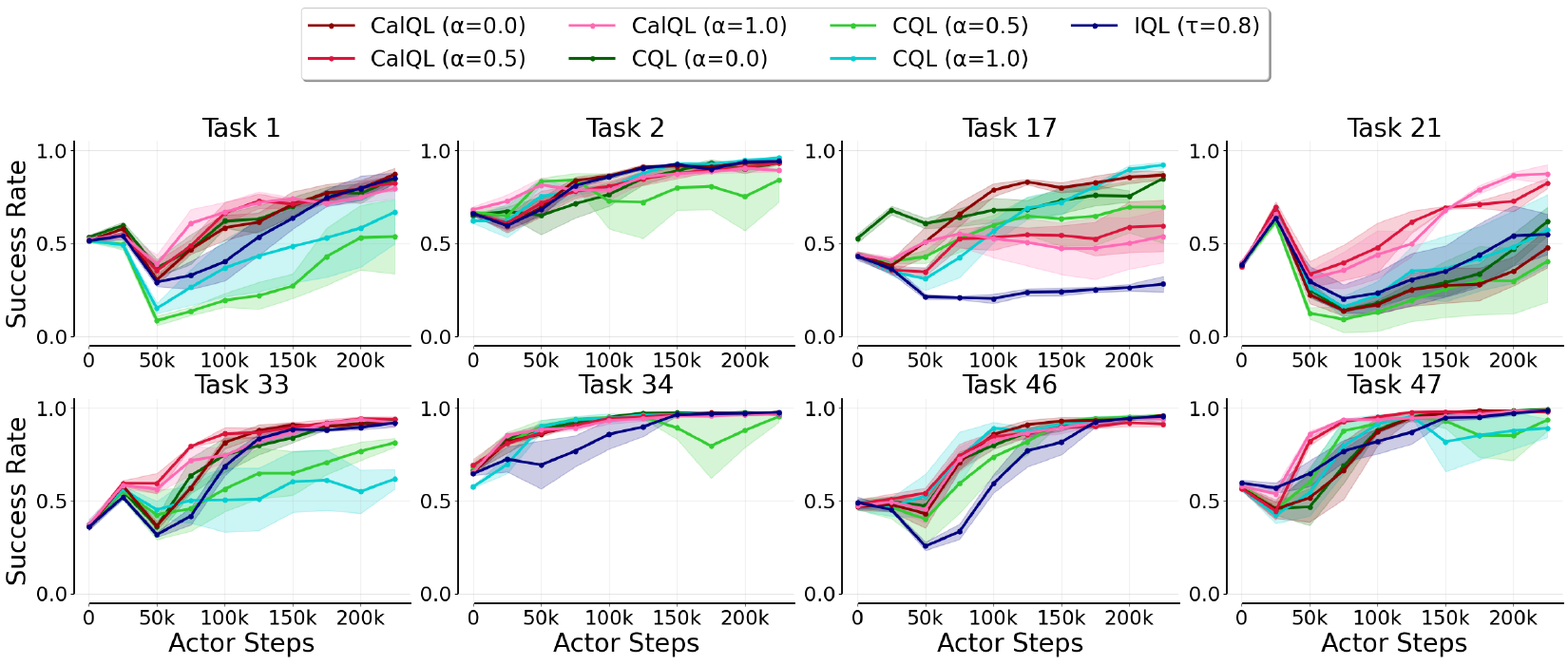}
    \vspace{-20mm}
    \caption{\textbf{Offline pre-training ablation. } Mean and 95\% CIs of rollout performance across 3 seeds.}
    \label{fig: exp_ablation_diff_q}
\end{figure}

\paragraph{Critic pre-training}
While warm-start through pre-training the critic is beneficial to asymptotic performance and prevents initial performance drop, the careful selection of the pre-training method could be important as well. We compare using CQL, Cal-QL, and IQL to the pre-training method. We consider using only 50 trajectories of successful trials of the base policy, while the standard offline RL benchmark tends to have far larger data volume~\citep{fu2020d4rl}. In \Cref{fig: exp_ablation_diff_q}, online performance using the Cal-QL pre-trained critic is consistently better and is robust to the conservative coefficients \(\alpha\). CQL demonstrates the worst performance with a severe forgetting issue, which aligns with the previous study~\citep{cal-QL}.

\begin{figure}
    \centering
    \includegraphics[width=0.85\linewidth]{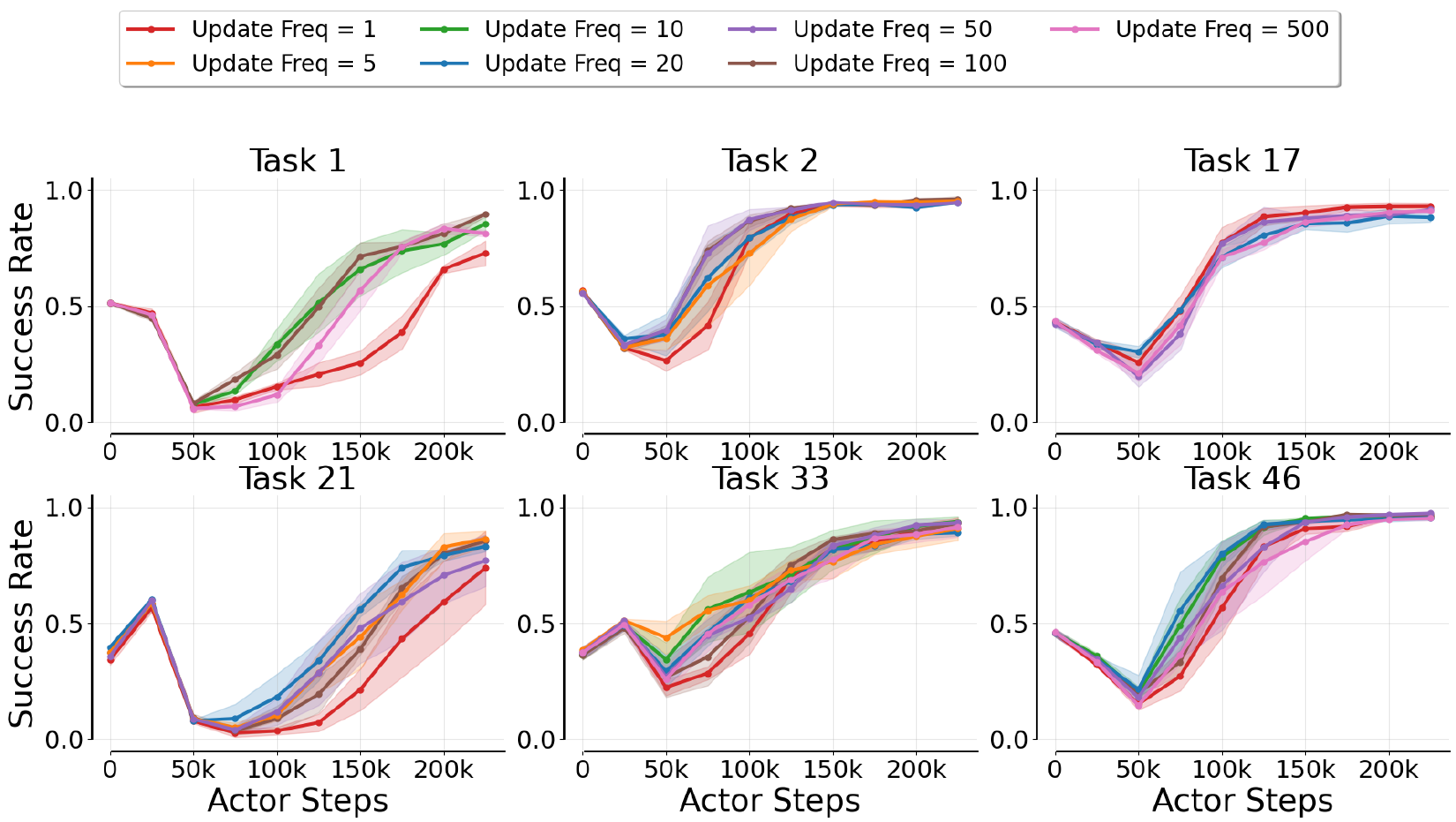}
    \caption{\textbf{Update frequency ablation.} Mean and 95\% CIs of rollout performance across 3 seeds.}
    \label{fig:rl-ablation-update-freq}
\end{figure}

\paragraph{Update frequency}
In the SERL pipeline, data collection and policy learning run asynchronously and periodically exchange network parameters and online data. We ablate the \emph{update frequency}—the number of gradient steps performed by the learner between parameter synchronizations with the data-collection actor—sweeping from 1 to 500. As shown in~\Cref{fig:rl-ablation-update-freq}, overall performance is largely insensitive to this hyperparameter, indicating robustness across a wide range of synchronization cadences.

\begin{figure}
    \centering
    \includegraphics[width=0.85\linewidth]{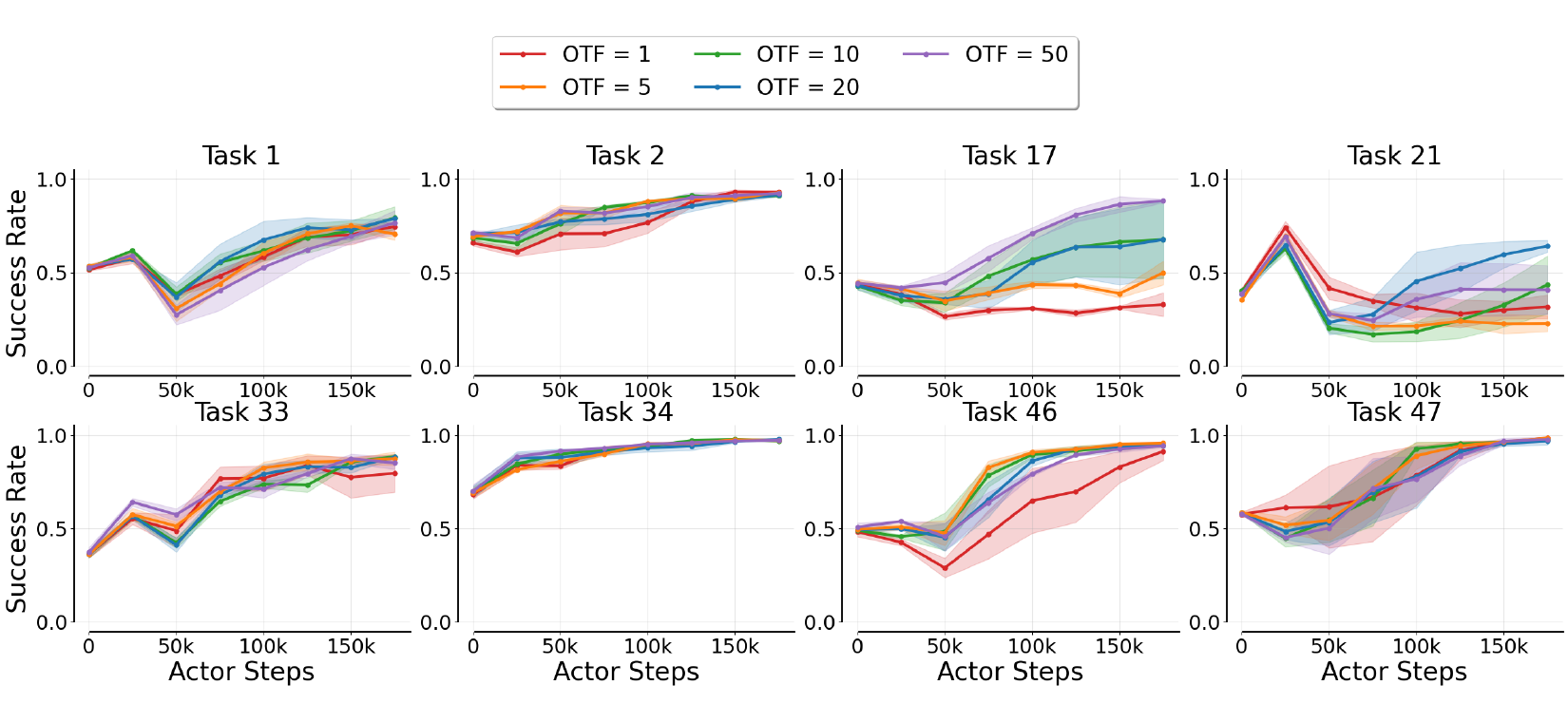}
    \caption{\textbf{On-the-fly Policy Ablation.} Mean and 95\% CIs of rollout performance across 3 seeds.}
    \label{fig:on-the-fly}
\end{figure}

\paragraph{On-the-Fly Policy}
On-the-fly (OTF) policy is introduced in~\citep{dong2025expo} to more effectively maximize the value function. It samples multiple actions and backs up the maximum Q value during TD learning. We adopt OTF to \method while only sampling multiple actions from the residual policy $\pi_\delta$ and conditioned on a fixed base action. We compare different sample sizes in~\Cref{fig:on-the-fly}. We found that OTF can improve sample efficiency, and a larger sample size ($> 20$) shows significant performance gain. But empirically, the asymptotic performance will eventually be similar. We use OTF$=1$ by default.

\begin{figure}
    \centering
    \includegraphics[width=0.85\linewidth]{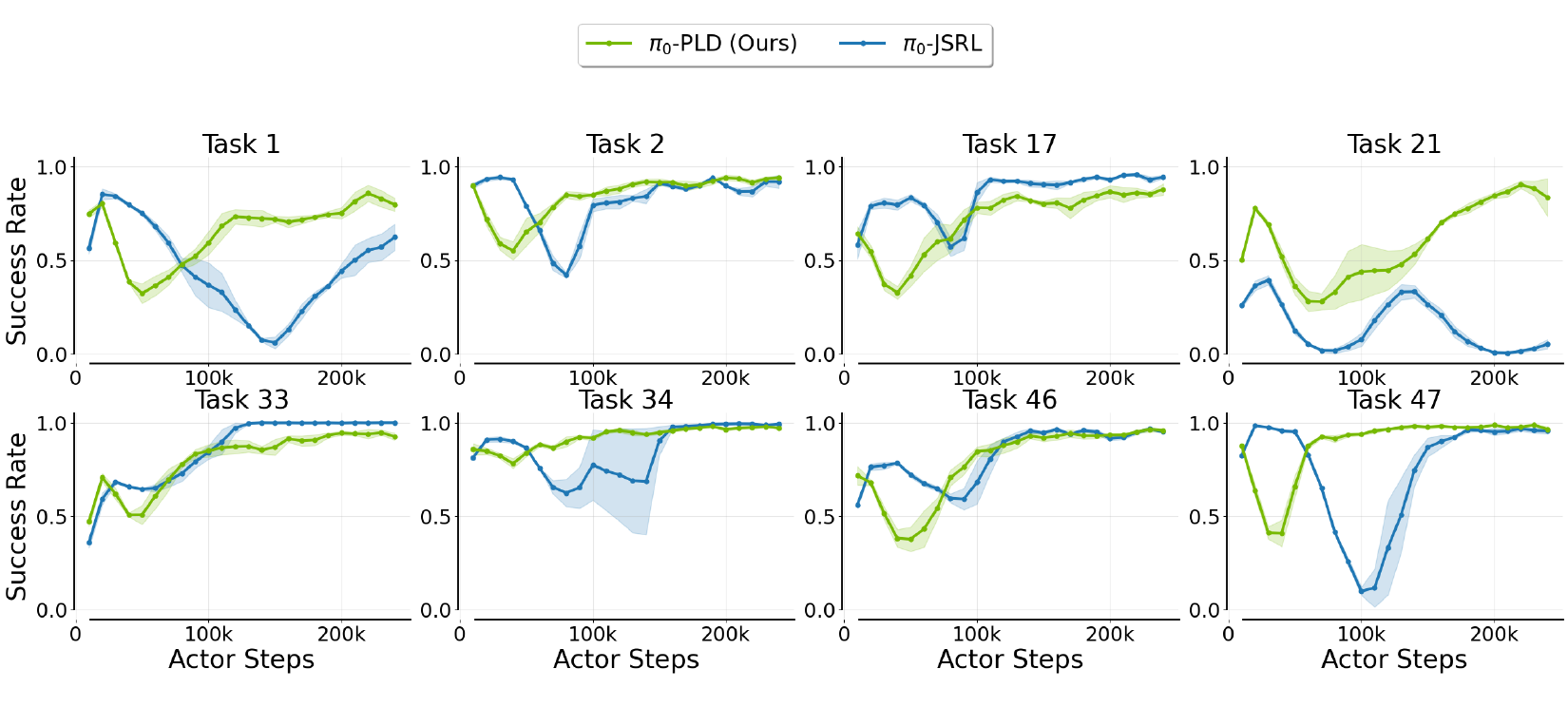}
    \caption{\textbf{Compared with JSRL.} Mean and 95\% CIs of rollout performance across 3 seeds.}
    \label{fig:jsrl}
\end{figure}

\paragraph{JSRL}
We further provide results, including JSRL~\cite{JSRL} in~\Cref{fig:jsrl}. We modify the original implementation by opting for a linear scheduler. JSRL demonstrates high data efficiency in general, but could fail to converge on some tasks. While \method can reliably provide solutions for all tasks.

\section{Implementation Details}
\subsection{RL Algorithm}
To ensure apples-to-apples comparisons, all baselines in~\Cref{subsec: RL exp} and~\Cref{sec: RL design} use the same network architecture—an \emph{3-layer MLP Gaussian policy} and \emph{Clipped Double Q-networks (CDQ)}~\cite{fujimoto2018td3} with \emph{LayerNorm}~\citep{ba2016ln}. Both actor and critic use a pre-trained ResNetV1-10 encoder to extract visual information. We present a detailed hyperparameter setting in~\Cref{tab: hyperparams}.

\begin{table}[h]
\caption{RL hyperparameter settings. We share the same setting across all tasks.}
\label{tab: hyperparams}
\vskip 0.15in
\begin{center}
\begin{small}
\begin{tabular}{l|c}
\toprule
\textbf{Hyperparameter}       & \textbf{Value}   \\ 
\midrule
\multicolumn{2}{l}{\textbf{Training}} \\ \midrule
Batch size                    & 256 \\
Buffer capacity               & 250000  \\
Discount factor ($\gamma$)    & 0.99    \\
Gradient clipping norm        & 1.0     \\
Learning rate                 & $3 \times 10^{-4}$  \\
Optimizer                     & AdamW  \\
Reward bias                   & 0.0  \\
Warmup episodes               & 100  \\
Critic to actor ratio         & 2  \\
On-the-fly ratio              & 1  \\

\midrule
\multicolumn{2}{l}{\textbf{Residual Policy}} \\ \midrule
Target entropy                & \(-\frac{act\_dim}{2}\)  \\
Initial temperature ($\tau$)  & 1.0 \\    
Action scale ($\xi$)          & 0.5 \\

\midrule
\multicolumn{2}{l}{\textbf{Critic}} \\ \midrule
Q functions ensemble           & 2   \\
Target update rate            & 0.005   \\

\midrule
\multicolumn{2}{l}{\textbf{Architecture}} \\ \midrule
Visual Encoder                & ResNetv1-10 \\
Hidden layer dimension        & 256 \\
Latent space dimension        & 256 \\
Q function dropout           & 0.0 \\
Activation                    & Tanh \\
Normalization                 & LayerNorm \\
\bottomrule
\end{tabular}
\end{small}
\end{center}
\vskip -0.1in
\end{table}

\subsection{SFT}
For fine-tuning either OpenVLA or $\pi_0$, we employ 8 $\times$ NVIDIA L40 GPU for LoRA~\citep{hu2021loralowrankadaptationlarge} fine-tuning with rank 32. For both $\pi_0$, and OpenVLA-OFT, we use the default hyperparameters from their open-source codebase.

\section{Real-world Experiments}
\label{sec:appendix-real-world-exp}

\subsection{Experiment Setup}
\label{sec:real-world-setup}
We deploy \method on a 7-DoF Franka Emika Panda with end-effector delta pose control at 20~Hz. The robot is equipped with one wrist-mounted camera, one side-view camera, and proprioceptive sensing as inputs. For each task, we pretrain an independent binary reward classifier by collecting a small-scale dataset of success and failure states. The model structure follows the setup in~\citep{luo2025hilserl}, which use a pretrained ResNet-10 and a 3-layer MLP model. We ensure the trained classifier using augmented false positive samples until it achieves 99\% success rate for each task. Due to the 3D printed desk, we don't need to reset the environment for the pick-cube task. \method performs auto-reset, residual RL training, and SFT automatically without human supervision. For the peg-insertion task (depicted in~\Cref{fig:real-world-generalization}), human supervisors need to randomly move the position of the hole to increase diversity.

\subsection{Generalization Performance}
\label{sec: generalization}
We perform SFT of $\pi_0$ on \textit{Pick Up Blue Cube (Clean Env)} and \textit{Peg Insertion} data, and evaluate the fine-tuned policy on \textit{Pick Up Blue Cube (Cluttered Env)} and \textit{Pick Up Red Cube (Cluttered Env)} tasks. The results in~\Cref{tab:real-world-generalize} show that VLA SFT on \method data achieves better generalization performance compared to human teleoperation data.

\begin{table}[h]
\centering
\caption{Comparison of PLD vs Human Data on real-world unseen tasks (success rate).}
\label{tab:real-world-generalize}
\setlength{\tabcolsep}{6pt}
\begin{tabular}{lcc}
\toprule
\textbf{SR \ Dataset} & \textbf{PLD Data} & \textbf{Human Data} \\
\midrule
Pick Up Blue Cube (cluttered Env) & 28/30 (93.3\%) & 12/30 (40.0\%) \\
Pick Up Red Cube (cluttered Env) & 20/30 (66.7\%) & 10/30 (33.3\%) \\
Peg Insertion & 30/30 (100.0\%) & 30/30 (100.0\%) \\
\bottomrule
\end{tabular}
\end{table}

%% file: asset/tables/task_description.tex
\begin{center}
\begin{longtable}{@{} 
    >{\rule{0pt}{4\baselineskip}\centering\arraybackslash}m{0.19\textwidth}   
    >{\rule{0pt}{4\baselineskip}}p{0.75\textwidth}                             
  @{}}
  \caption{Selected LIBERO-90 Tasks for RL Ablation Studies.}
  \label{tab:task_details}\\

  \toprule
  \multicolumn{1}{p{0.19\textwidth}}{\centering\textbf{Visualization}}
    & \multicolumn{1}{p{0.75\textwidth}}{\centering\textbf{Task Name and Description}} \\
  \midrule
  \endfirsthead

  \multicolumn{2}{@{}l}{\small\itshape continued from previous page}\\
  \toprule
  \multicolumn{1}{p{0.19\textwidth}}{\centering\textbf{Visualization}}
    & \multicolumn{1}{p{0.75\textwidth}}{\centering\textbf{Task Name and Description}} \\
  \midrule
  \endhead

  \midrule
  \multicolumn{2}{r}{\small\itshape continued on next page} \\
  \endfoot

  \bottomrule
  \endlastfoot

  \includegraphics[width=\linewidth]{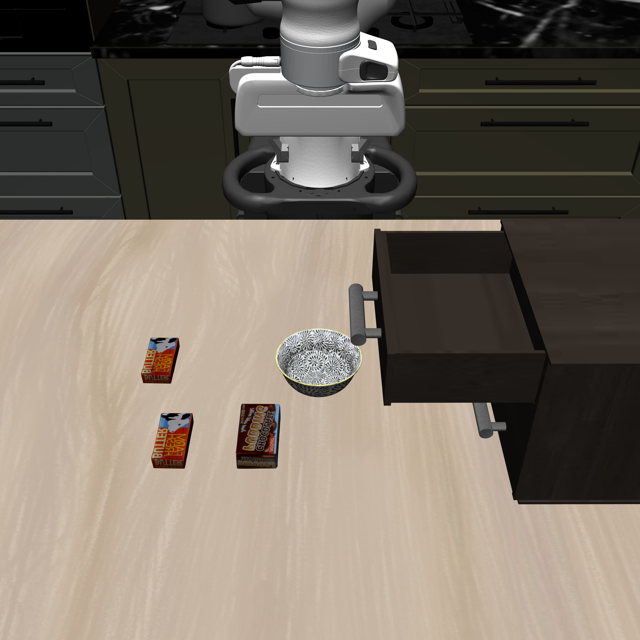}
    & \parbox[c][4\baselineskip][c]{\linewidth}{\textbf{LIBERO Task 1}\\ close the top drawer of the cabinet and put the black bowl on top of it}\\
  \midrule

  \includegraphics[width=\linewidth]{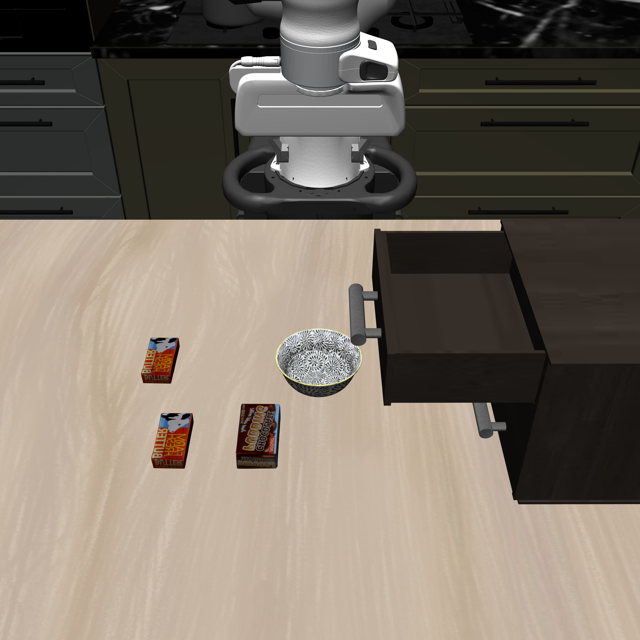}
    & \parbox[c][4\baselineskip][c]{\linewidth}{%
        \textbf{LIBERO Task 2}\\ put the black bowl in the top drawer of the cabinet
      } \\
  \midrule

  \includegraphics[width=\linewidth]{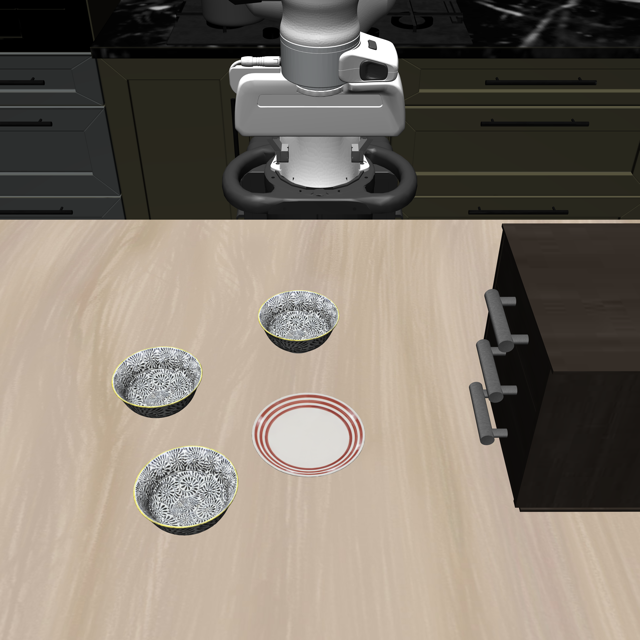}
    & \parbox[c][4\baselineskip][c]{\linewidth}{%
        \textbf{LIBERO Task 17}\\
        stack the middle black bowl on the back black bowl
      } \\
  \midrule

  \includegraphics[width=\linewidth]{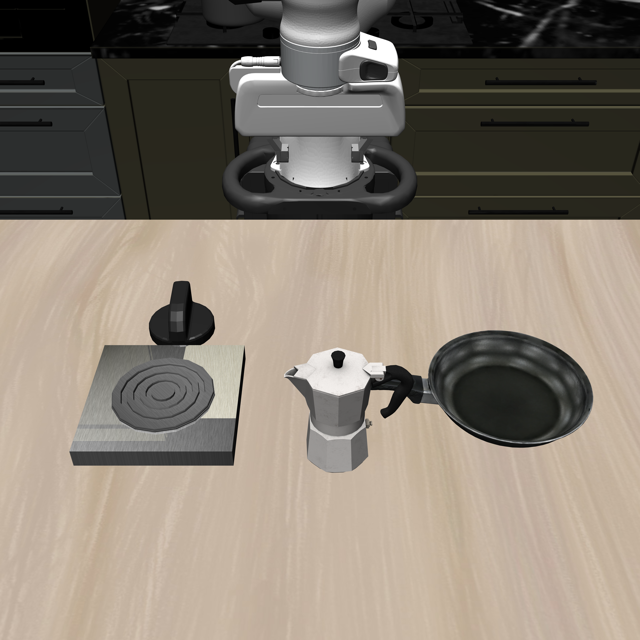}
    & \parbox[c][4\baselineskip][c]{\linewidth}{%
        \textbf{LIBERO Task 21}\\
        turn on the stove and put the frying pan on it
      } \\
  \midrule

  \includegraphics[width=\linewidth]{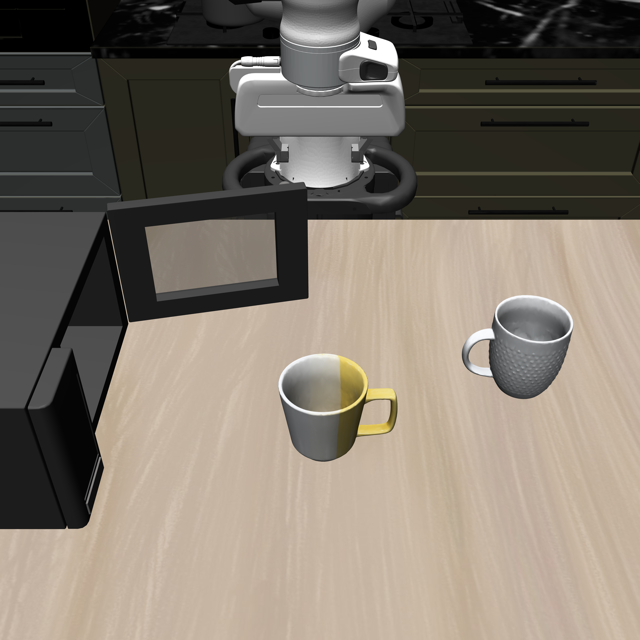}
    & \parbox[c][4\baselineskip][c]{\linewidth}{%
        \textbf{LIBERO Task 33}\\
        close the microwave
      } \\
  \midrule

  \includegraphics[width=\linewidth]{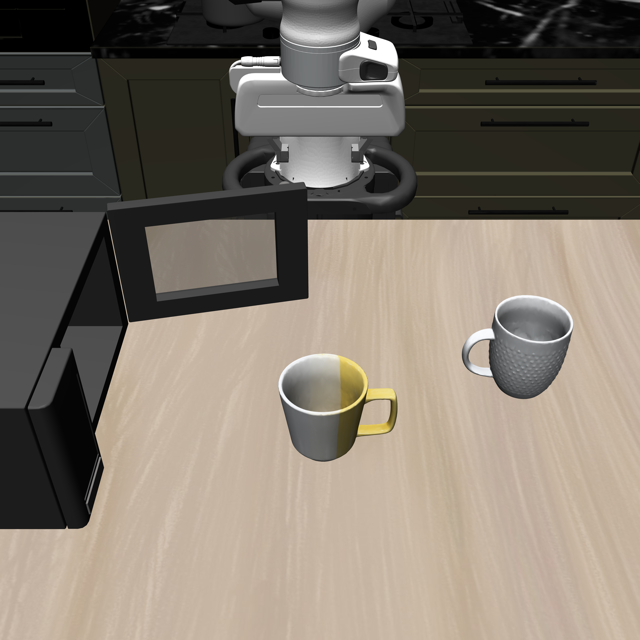}
    & \parbox[c][4\baselineskip][c]{\linewidth}{%
        \textbf{LIBERO Task 34}\\
        put the yellow and white mug to the front of the white mug
      } \\
  \midrule

  \includegraphics[width=\linewidth]{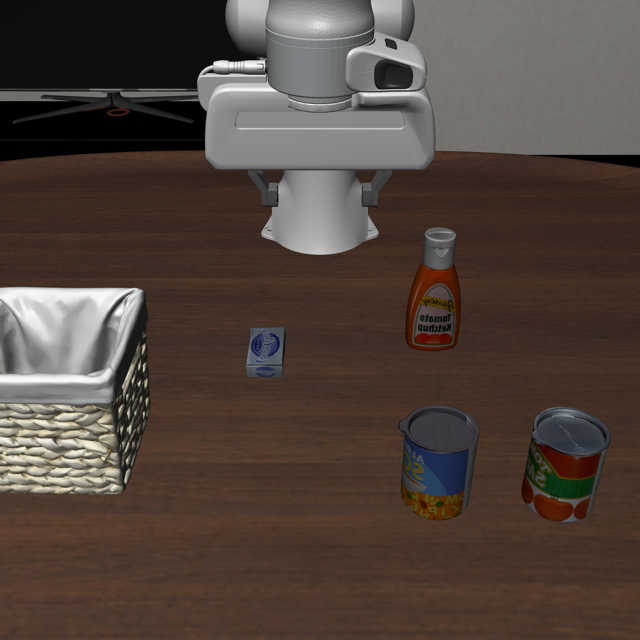}
    & \parbox[c][4\baselineskip][c]{\linewidth}{%
        \textbf{LIBERO Task 46}\\
        pick up the alphabet soup and put it in the basket
      } \\
  \midrule

  \includegraphics[width=\linewidth]{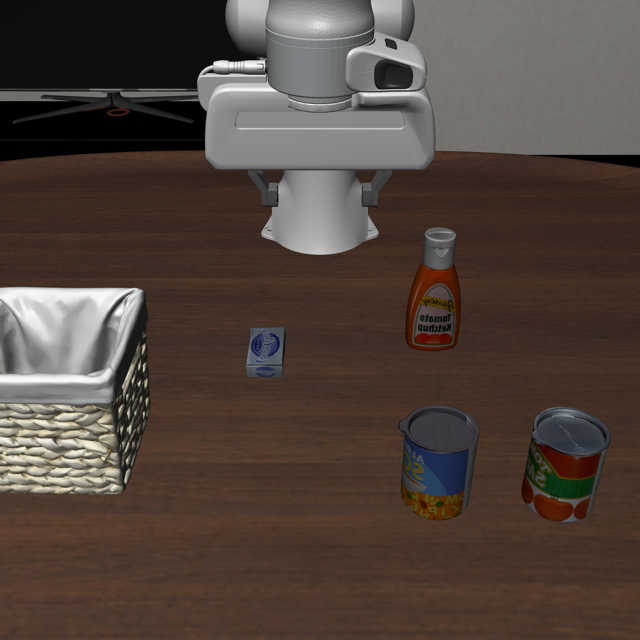}
    & \parbox[c][4\baselineskip][c]{\linewidth}{%
        \textbf{LIBERO Task 47}\\
        pick up the cream cheese box and put it in the basket
      } \\
  \midrule

\end{longtable}
\end{center}